\def\year{2021}\relax
\documentclass[letterpaper]{article} % DO NOT CHANGE THIS
\usepackage{aaai21}  % DO NOT CHANGE THIS
\usepackage{times}  % DO NOT CHANGE THIS
\usepackage{helvet} % DO NOT CHANGE THIS
\usepackage{courier}  % DO NOT CHANGE THIS
\usepackage[hyphens]{url}  % DO NOT CHANGE THIS
\usepackage{graphicx} % DO NOT CHANGE THIS
\usepackage{natbib}  % DO NOT CHANGE THIS AND DO NOT ADD ANY OPTIONS TO IT
\usepackage{caption} % DO NOT CHANGE THIS AND DO NOT ADD ANY OPTIONS TO IT
\nocopyright
\usepackage{microtype}
\usepackage{graphicx}
\usepackage{amsmath}
\usepackage{lipsum}
\usepackage{algorithmicx}

\usepackage{subcaption}
\usepackage{comment}
\usepackage{url}
\usepackage{multirow}
\usepackage{multirow}
\usepackage{booktabs}
\DeclareUnicodeCharacter{2212}{-}

\title{Analysis of Twitter Users' Lifestyle Choices using Joint Embedding Model}
\author {
    Tunazzina Islam, 
    Dan Goldwasser  \\
}
\affiliations {
    Department of Computer Science, Purdue University\\
    West Lafayette, Indiana 47907 \\
    islam32@purdue.edu, dgoldwas@purdue.edu
}

\begin{document}

\maketitle

\begin{abstract}
Multiview representation learning of data can help construct coherent and contextualized users' representations on social media. This paper suggests a joint embedding model, incorporating users' social and textual information to learn contextualized user representations used for understanding their lifestyle choices. We apply our model to tweets related to two lifestyle activities, \textit{`Yoga' and `Keto diet'} and use it to analyze users' activity type and motivation. We explain the data collection and annotation process in detail and provide an in-depth analysis of users from different classes based on their Twitter content. Our experiments show that our model results in performance improvements in both domains.  
\end{abstract}
\section{Introduction}
Nowadays, people express opinions, interact with friends and share ideas and thoughts via social media platforms. The data collected by these platforms provide a largely untapped resource for understanding lifestyle choices, health, and well-being \cite{islam2019yoga,amir2017quantifying,schwartz2016predicting,yang2016life,schwartz2013characterizing}. 

In this paper, we use Twitter to study two lifestyle-related activities, Yoga -- a popular multi-faceted activity and Ketogenic diet (often abbreviated as Keto) -- a low-carbohydrate, high-fat, adequate-protein diet. 
Various studies show that yoga offers physical and mental health benefits for people of all ages \cite{ross2010health,smith2009evidence,yurtkuran2007modified,khalsa2004treatment}.
Ketogenic diet recently discovered benefits include weight loss \cite{johnstone2008effects}, reversal/control of type 2 diabetes \cite{mckenzie2017novel} as well as therapeutic potential in many pathological conditions, such as polycystic ovary syndrome, acne, neurological diseases, cancer, and the amelioration of respiratory and cardiovascular disease risk factors \cite{paoli2013beyond}.

The goal of this paper is to analyze the different lifestyle choices of users based on their tweets.  These users can correspond to practitioners who share their journey and explain their motivation when taking a specific lifestyle, but also to commercial parties and interest groups that use social media platforms to advance their interests.
%
% Interests in these topics can come from different types of users based on different motivations.
%
% A ketogenic diet could be an interesting alternative to treat certain conditions and may accelerate weight loss. 
\begin{figure}[htbp]
  \centering  
  \includegraphics[width= 0.46 \textwidth]{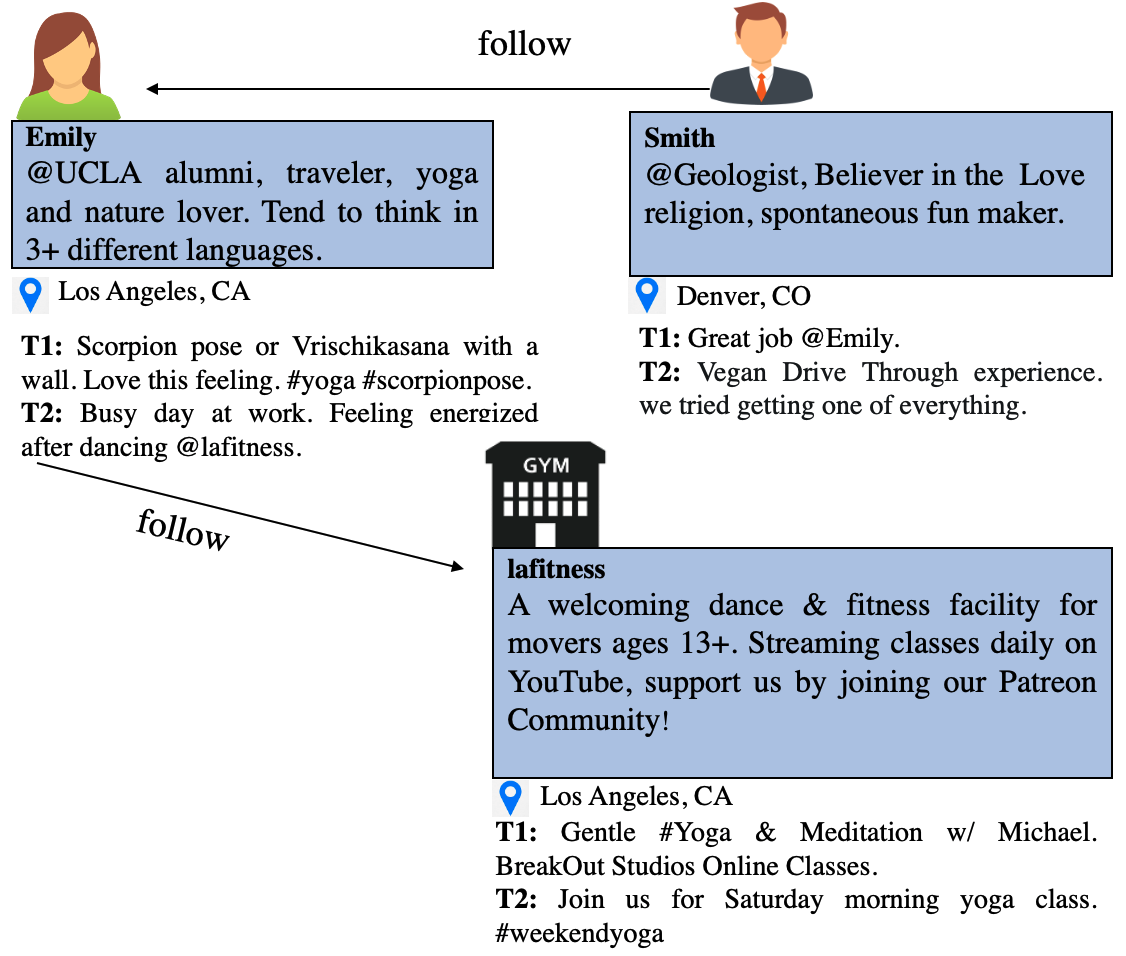}
    \caption{User and Network information on Twitter. Blue rectangle boxes are user profile description and $T1, T2$ are tweets of corresponding users.}
    \label{fig:twt_net}
\end{figure}
Because of the short and often ambiguous nature of tweets, a simple pattern-based analysis using yoga-related keywords often falls short of capturing relevant information. Fig. \ref{fig:twt_net} shows three different Twitter users, the same ``\#yoga” is used by two different types of users. Our main insight in this paper is that understanding user types and their motivation should be done collectively over the tweets, profile information, and social behavior. In the above example, 
% By looking at their description, location, tweets, and network information, we can infer users' activity type and motivation. 
\textbf{Emily}, a practitioner, tweets about specific yoga poses and the emotions the activity evokes. The user \textbf{lafitness}, a gym, tweets about online yoga classes in their studios. In addition to the tweets' contents, the profile description of \textbf{Emily} indicates that she is a practitioner. And on the other hand, the profile description of \textbf{lafitness} indicates that it is a promotional account (Fig. \ref{fig:twt_net}). Moreover, social information can help to further disambiguate the text based on the principle of homophily \cite{mcpherson2001birds}, the user types and motivations are likely to be reflected by their social circles.

\begin{figure*}[htbp]
  \centering  
  \includegraphics[width= 0.95\textwidth]{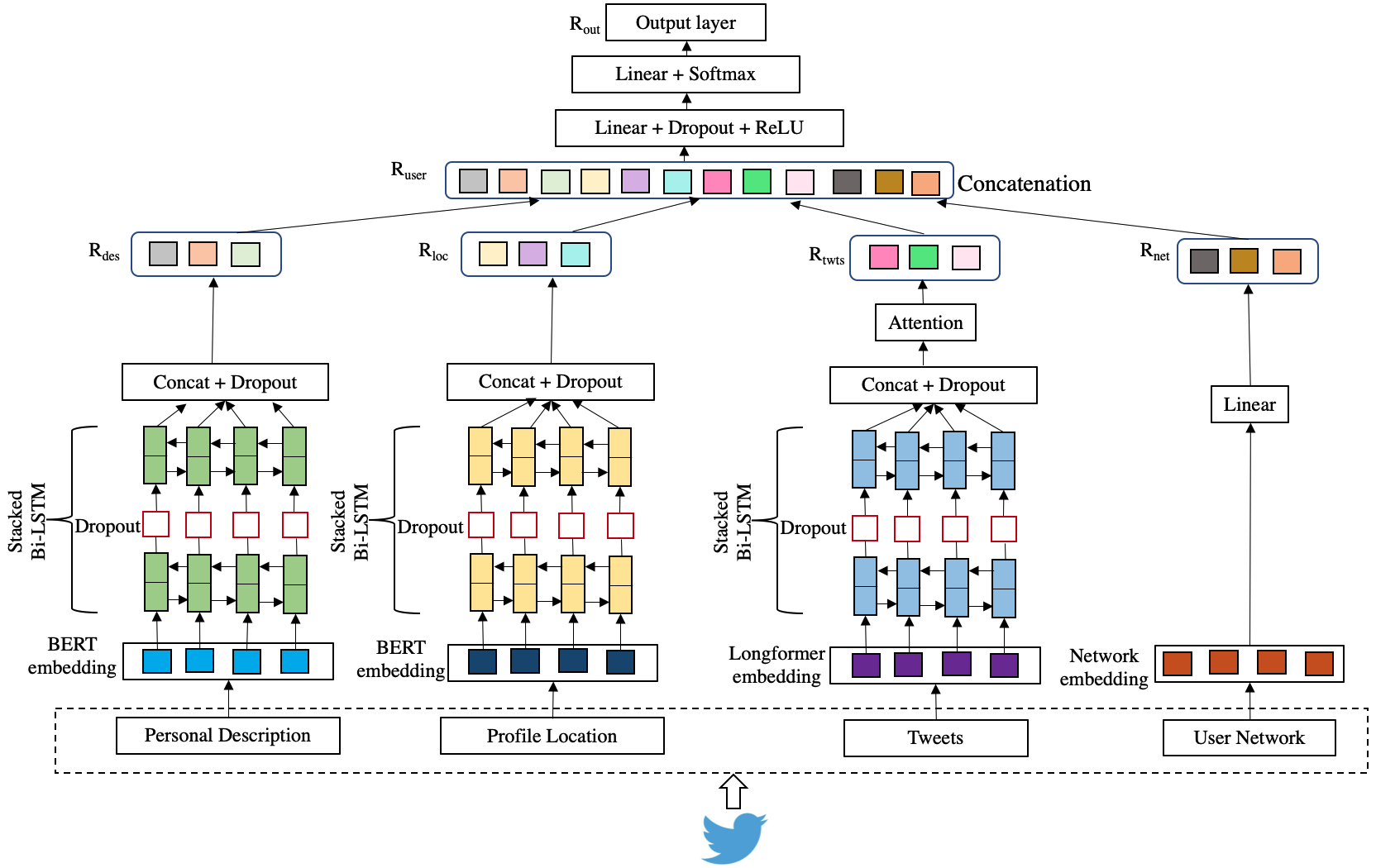}
    \caption{Architecture of our model.}
    \label{fig:model}
\end{figure*}

Past work aiming to understand Twitter users' demographic properties has also used a combination of their tweets, and social information \cite{li2015learning,benton2016learning,yang2017overcoming,mishra2018neural,del2019you}, however unlike these works, which look at general demographic properties, our challenge is to construct a user representation relevant for characterizing nuanced, activity and lifestyle specific properties.
Recently, pre-trained contextualized language models,
such as ELMo \cite{peters2018deep}, OpenAI GPT
\cite{radford2018improving}, and BERT \cite{devlin2019bert}, have led to significant improvements in several NLP tasks. However, not much improvement on our task was obtained by directly using the pre-trained BERT model, as it falls short of representing non-linguistic information. 

We suggest a method for combining large amounts of Twitter content and social information associated with each user. 
We concatenate all yoga-related (for keto diet, all keto-related) tweets associated with a given user and use pre-trained Longformer \cite{beltagy2020longformer} -- a BERT-like model for long documents, for the contextualized embedding of the user tweets. Many users share their profile description and their location on Twitter; we add this information to our model using a pre-trained BERT model. We refer to our model as BERT based joint embedding model. Finally, we embed the social information associated with the Twitter users and concatenate the user's embedded representation to their profile description and content representations.

Using this model, we predict (i) the user type, i.e., whether they are a practitioner, a promotional user, or other (ii) the user's motivation, e.g., practicing yoga for health benefits, spiritual growth, or other reasons such as commercially motivated users. We compare our model's performance against several other modeling choices. We describe the data collection and annotation details as well as the results of predicting user type and motivation using our model and in-depth analysis of users from different classes. The main contributions of this work are as follows:

\begin{enumerate}
    \item Formulating an information extraction type for lifestyle activities, characterizing activity-specific user types, and motivations. We create an annotated dataset related to `yoga' and `keto diet'.
    \item We suggest a model for aggregating users' tweets as well as metadata and contextualizing this textual content with social information.
   
    \item We perform extensive experiments to empirically evaluate the contribution of different types of information to the final prediction, and we show that their combination results in the best performing model.

    % \item Extensive empirical evaluation, comparing twelve different settings: (i) Description; (ii) Location; (iii) Tweets; (iv) Network; (v) BERT fine-tuned with Description (Des\_BF); (vi) BERT fine-tuned with Location (Loc\_BF); (vii) BERT fine-tuned with Tweets (Twts\_BF); (viii) joint embedding on description and location (Des + Loc); (ix) joint embedding on description and network (Des + Net); (x) joint embedding on description, location, and tweets (Des + Loc + Twt); (xi) joint embedding on description, location, and network (Des + Loc + Net), (xii) Word2Vec based joint embedding on description, location, tweets, and network. We show that our Bert based joint embedding model outperforms these settings.
    \item We conduct a qualitative analysis aimed at describing the relationship between the output labels and several different indicators, including the tweets, profile descriptions, and location information.
\end{enumerate}

% \textbf{Reproducibility:}
% Our code and public data are available \href{https://github.com/tunazislam/Joint-Embedding-Model}{here}.\footnote[1]{\url{https://github.com/tunazislam/Joint-Embedding-Model}}
The rest of the paper is organized as follows:
we start with the discussion of related work;
next, we provide the problem definition of our work; then, describe the technical approach; next, dataset and annotation details; later, we elaborate details of experimental settings, including discussion of the baseline models, hyperparameter tuning; finally, we show results and analysis containing ablation study, error analysis. Our code and public data are available here.\footnote[1]{\url{https://github.com/tunazislam/Joint-Embedding-Model}}

\section{Related Work}
\label{sec:2}
Prior research has demonstrated that we can infer many latent user characteristics by analyzing the information in a user's social media account, i.e., personality \cite{kosinski2013private,schwartz2013personality}, emotions \cite{wang2015detecting}, happiness \cite{islam2020does}, mental health \cite{amir2017quantifying}, mental disorders \cite{de2013predicting,reece2017forecasting}. \cite{mishra2018neural,mishra2019abusive} exploited user's community information along with textual features for detecting abusive instances. \cite{ribeiro2018characterizing} characterized hateful users using content as well as user’s activity and connections. \cite{miura2017unifying,ebrahimi2018unified,huang2019hierarchical} used a joint model incorporating different types of available information including tweet text, user network, and metadata for geolocating Twitter users. 
%Recently \cite{islam2021you,islam2020does} show joint embedding model to understand users’ types and motivations using social and textual information. 
In contrast, our approach relies on contextualized embeddings for user metadata and tweets. In this paper, we leverage Twitter content and social information associated with each user to learn user representation for predicting the following three tasks: (i) yoga user type, (ii) user motivation related to yoga, (iii) keto user type.

\section{Problem Definition}
We formulate our problem as multiview representation neural network based fusion. Suppose we have two learned feature maps $X_a$ and $X_b$ for views $a$ and $b$, where weights are shared across two views. Concatenation fusion is as follows:

\begin{align}
h_{cat} = [x_a, x_b] \label{eq:1}
\end{align}
where data from multiple views are integrated into a single
representation $h$, which exploits the knowledge
from multiple views to represent the data. 

\section{Methodology}
\label{sec:4}
We use the following sources of information to train our model: 1) Tweet text; 2) User network; and 3) Metadata, including user location and description. Our model employs those sources and then jointly builds a neural network model to generate a dense vector representation for each field and finally concatenates these representations. 
Fig. \ref{fig:model} shows the overall architecture of our proposed model.

% \subsection{Text Component}
% User description, location, tweets contain text. Let's say, $des = \{wd_1, wd_2, ..., wd_n\}$ denotes profile description of a user $u$ where, $n$ is the number of words in the description, and $loc = \{wl_1, wl_2, ..., wl_p\}$ is the user $u$'s location where, $p$ is the number of words in location. The user $u$ generates multiple tweets $T_1 = \{wt_{11}, wt_{12}, ..., wt_{1n}\}, ..., T_m = \{wt_{m1}, wt_{m2}, ..., wt_{mn}\}$, where $m$ is the total tweets of the user. For user tweets, we concatenate all yoga related tweets which represents a long document $twts = \{T_1||...||T_m\}$. We denote the concatenation operation as $||$.

\subsection{Metadata Representation} 
The metadata embedding transforms each metadata into a fix-sized embedding vector. In this paper, we focus on two metadata fields: user description and location. We use pre-trained uncased BERTbase model. Transformers \cite{vaswani2017attention} in BERT consist of multiple layers, each of which implements a self-attention
sub-layer with multiple attention heads. We pass metadata embedding to stacked Bi-LSTM \cite{hochreiter1997long}. We get the final hidden representation of metadata by concatenating the forward and backward directions. $R_{des}$ and $R_{loc}$ are the representations of user description and location respectively.

\subsection{Tweet Representation}
For user tweets, we concatenate all yoga-related tweets, which represents a long document. Similarly, for keto, we concatenate all keto-related tweets. Then we use pre-trained longformer-base-4096 model started from the RoBERTa \cite{liu2019roberta} checkpoint and pre-trained on long documents. Longformer uses a combination of a sliding window (local) attention and global attention. We forward the tweet embedding to stacked Bi-LSTM. We get the hidden representation of tweets by concatenating the forward and backward directions. To assign important words in the final representation ($R_{twts}$), we use a context-aware attention mechanism \cite{bahdanau2015neural}. %$R_{twts}$ is the user tweet representation.

\subsection{User Network Representation}
\label{subsec:4.3}
To build a dense user network, we consider those users from our dataset if they are $@-$mentioned \cite{rahimi2015exploiting} in other users' (from our data) tweets. We create an undirected and unweighted graph from interactions among users via retweets and/or $@-$mentions. Nodes are all users in our dataset. An edge is created between two users if either user mentions the other (from our data). In this work, we do not consider edge weights. For yoga, we have $534$ nodes with $1831$ edges, and for the keto diet, there are $234$ nodes with $809$ edges. To compute node embedding, we use Node2Vec \cite{grover2016node2vec}. For every node $u$, Node2Vec's mapping function maps $u$ to a low dimensional embedding of size $d$ that maximizes the probability of observing nodes belonging to $S(u)$ which is the set of $n$ nodes contained in the graph by taking $k$ random walks starting from $u$. We generate the embedding of user network $E_{net} = (e_{u_1},..., e_{u_V})$ and forward to a linear layer to compute user network representation, $R_{net}$.

% For every node $u$, Node2Vec creates a mapping function $f : V \longrightarrow \mathbb{R}^d$ which maps $u$ to a low dimensional embedding of size $d$ that maximizes the probability of observing nodes belonging to $S(u)$ which is the set of $n$ nodes contained in the graph by taking $k$ random walks starting from $u$. We generate the embedding of user network $E_{net} = (e_{u_1},..., e_{u_V})$ and forward to a linear layer to compute user network representation, $R_{net}$.

\subsection{User Representation}
\label{subsec:4.4}
The final user representation, $R_{user}$ is built by concatenation of the four views generated from four sub-networks description, location, tweets, and user network respectively (Fig. \ref{fig:model}). We define $R_{user}$ as follows:

\begin{align}
% R_{user} = R_{des} || R_{loc} || R_{twts} || R_{net} \label{eq:2}
R_{user} = [R_{des}, R_{loc}, R_{twts}, R_{net}] \label{eq:2}
\end{align}
%We denote the concatenation operation as $||$. 
$R_{user}$ is passed through a fully connected two-layer classifier where the first linear layer with ReLU \cite{nair2010rectified} activation function. The final prediction $R_{out}$ is passed through a softmax activation function.
The risk of overfitting is handled by using
dropout \cite{srivastava2014dropout} between individual neural network layers. We use stochastic gradient descent over shuffled mini-batches with Adam \cite{kingma2014adam} and cross-entropy loss as the objective function for classification.

\begin{table*}
  \centering
    \begin{tabular}{lccccccccccc}
    \toprule
    \textbf{Model} & \textbf{lr}  & \textbf{opt} & \textbf{reg} & \textbf{batch}  & \textbf{hd}   & \textbf{lstm}  & \textbf{attn} & \textbf{ls} & \textbf{eut} & \textbf{eum}\\
    \midrule
    Description  & 1$e^{-3}$ & Adam & 0 & 32 & 300 & 2 & - & - & 6 & 5\\
    Location & 1$e^{-3}$ & Adam & 0 & 32 & 300 & 2  & - & - & 5 & 5\\
    Tweets  & 1$e^{-3}$ & Adam & 0 & 32 & 300 & 2 & 300 & - & 8 & 6 \\
     Network  & 1$e^{-3}$ & Adam  & 0 & 32 & 150 & - & - & - & 4 & 8\\
    Des\_BF & 2$e^{-5}$ & AdamW & .01 & 32 & - & - & - & - & 4 & 4  \\
    Loc\_BF  & 2$e^{-5}$ & AdamW & .01 & 32 & - & - & - & - & 2 & 2  \\
    Twts\_BF  & 2$e^{-5}$ & AdamW & .01 & 32 & - & - & - & - & 2 & 4  \\
    Des + Loc  & 1$e^{-3}$ & Adam & 0 & 32 & 300 & 2 & - & 600 & 5 & 6\\
    Des + Net  & 1$e^{-3}$ & Adam & 0 & 32 & 300 & 2 & - & 600 & 5 & 5\\
    Des + Loc + Twt  & 1$e^{-3}$ & Adam & 0 & 32 & 300 & 2 & 300 & 600 & 5 & 7 \\
    Des + Loc + Net  & 1$e^{-3}$ & Adam & 0 & 32 & 300 & 2 & - & 600 & 6 & 7 \\[.2em]
    \textbf{Our model}  & 1$e^{-3}$ & Adam & 0 & 32 & 300 & 2 & 300 & 600 & 7 & 6 \\
    \bottomrule
    \end{tabular}\\[.2em]
    {\small
    \begin{tabular}{rp{38.5em}}
    	lr & Learning rate. \\
    	opt & Optimizer. \\
   		reg & Weight decay ($L^2$ regularization). \\
    	batch & Batch size. \\
    	hd & Hidden dimension. \\
    	lstm & Number of LSTM layer as we use stacked Bi-LSTM. \\
    	attn & Attention vector size. \\
    	ls & Size of the first layer of two-layer classifier. \\
    	eut & Best result achieved at epochs for user type classification. \\
    	eum & Best result achieved at epochs for user motivation classification. \\
	\bottomrule
    \end{tabular}}
    \caption{Hyperparameter details of the models.}
  \label{tab:hyperparam}%
\end{table*}%

\section{Dataset}
We download tweets using Tweepy by Twitter streaming API sub-sequentially from May to November, 2019. For yoga, we collect $419608$ tweets related to yoga containing specific keywords : `yoga', `yogi', `yogalife', `yogalove', `yogainspiration', `yogachallenge', `yogaeverywhere', `yogaeveryday', `yogadaily', `yogaeverydamnday', `yogapractice', `yogapose', `yogalover', `yogajourney'. There are $297350$ different users among them $13589$ users have at least five yoga-related tweets in their timelines. For this work, we randomly pick $1298$ users and collect their timeline tweets. We have $3097678$ timeline tweets in total.

For ketogenic diet, we focus on several keywords i.e., `keto', `ketodiet', `ketogenic', `ketosis', `ketogenicdiet', `ketolife', `ketolifestyle', `ketogenicfood', `ketogenicfoodporn', `ketone', `ketogeniclifestyle', `ketogeniccommunity', `ketocommunity', `ketojourney' to extract $75048$ tweets from $38597$ different users. Among them $16446$ users have at least two keto-related tweets in their timelines. In this paper, we randomly pick $1300$ users and we have total $3253833$ timeline tweets.

To pre-process the text, we first convert them into lower case, remove URLs, smiley, emoji. To prepare the data for input to BERT and Longformer, we tokenize the text using BERT and RoBERTa’s wordpiece tokenizer.

\subsection{Data Annotation} 
% We annotate $1300$ users based on the intent of the tweets and observation of user description. Consider the following three tweets: 

% {\small \textbf{Tweet 1:} \textit{Learning some traditional yoga with my good friend. \#yoga \#yogaeverydamnday \#healthylife} 

% \textbf{Tweet 2:} \textit{Our mission at 532Yoga is pretty simple; great teachers, great classes and superbly happy students \#yoga} 

% \textbf{Tweet 3:} \textit{Yoga is more than fitness, it’s a mental and spiritual release. \#MondayMotivation \#Yoga} }

% The intention of Tweet 1 is yoga activity (learning yoga), Tweet 2 is more about promoting a yoga studio, Tweet 3 is also about yoga activity (practicing yoga on Monday). We annotate user of Tweet 1 and Tweet 3 as practitioner and user of Tweet 2 as promotional. In Tweet 1, user motivation is health benefit, in Tweet 2, promotional user has other motivation, and the motivation of the user of Tweet 3 is spiritual. 

%The annotating procedure in experimental settings is unclear to me. First, for each user, will all the tweets in her timeline be considered, or just a small set of tweets. Second, the annotation is done by one person or multiple people? If it is done by multiple people, how to deal with the situation that people gives different label for each user.
To annotate the data, for each user, we check both their profile description and timeline tweets. For the tweets, we consider only yoga/keto-related tweets from their timeline. 
We first look at the user profile description for user type, whether they explicitly mention practicing a specific lifestyle (i.e., yogi, ketosis); then, we look for the user timeline tweets. If the user tweets about the first-hand experience of practicing yoga (i.e., \textit{love practicing yoga early in the morning})/keto diet (i.e., \textit{lost $5lb$ in $1^{st}$ week of keto}), we annotate the user as a `practitioner'.  After looking at the description and tweets, if we observe that they are promoting a gym/studio (i.e., \textit{offering free online yoga class}), online shop (i.e., \textit{selling yoga mat}), app (i.e., \textit{sharing keto food recipe}), restaurant, community etc., rather than sharing their first-hand experience about a particular lifestyle, we annotate them as a `promotional' user. If we notice a user has all retweets in their timeline tweets related to yoga/keto (they might have an interest in a particular lifestyle), we annotate them as `others'.

For user motivation, we check for the user timeline tweets. If the user tweets about practicing yoga for health benefit (i.e., \textit{yoga heals my back pain}), we annotate the user motivation as `health'.  If the user tweets about practicing yoga for spiritual help (i.e., \textit{yoga gives me a spiritual wisdom path}), we annotate the user motivation as `spiritual'.  Otherwise, we annotate the motivation as `others'.

\begin{figure}
\begin{subfigure}{\columnwidth}
  \centering
  \includegraphics[width=\textwidth]{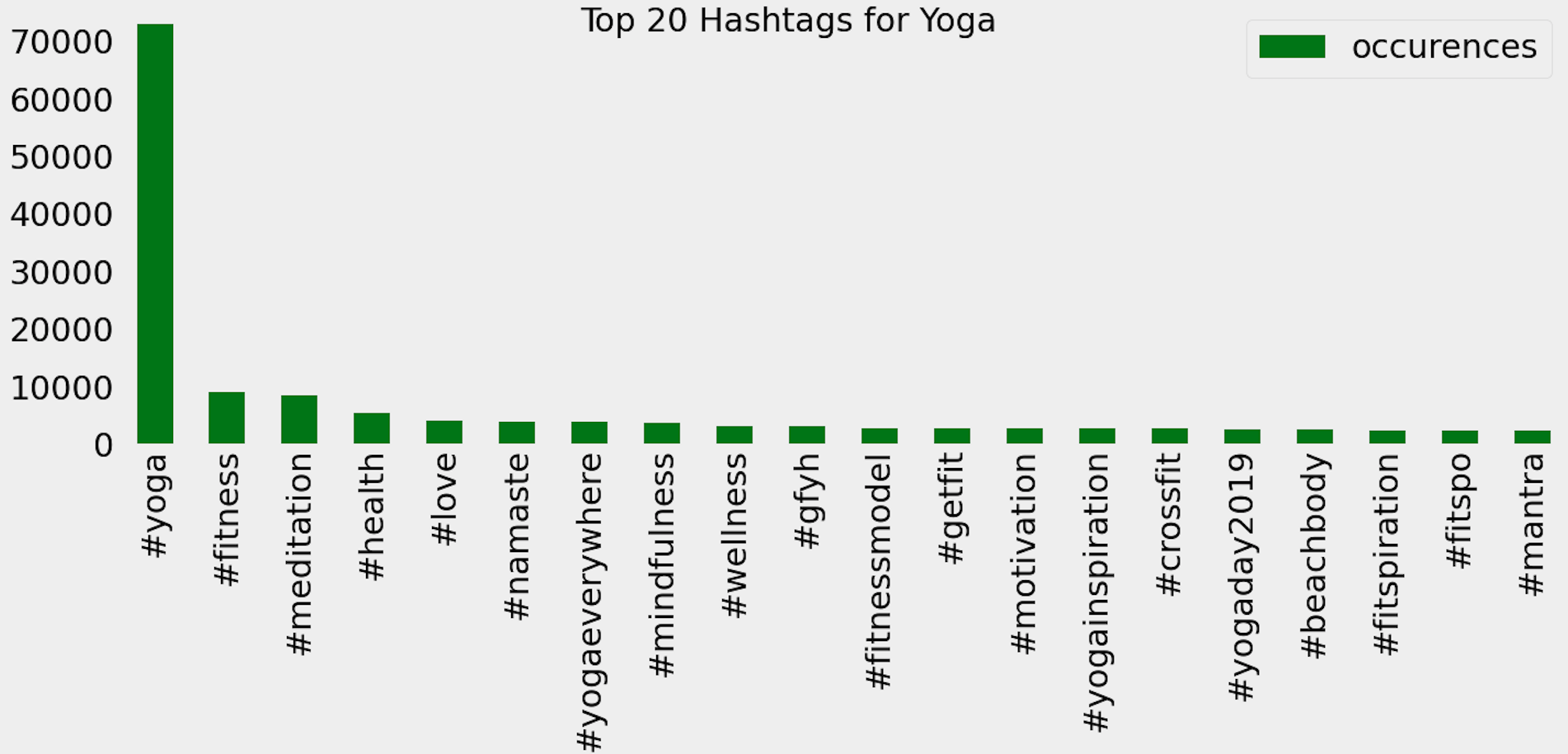}
  \caption{Yoga}
  \label{fig:hashtag_yoga_bar}
\end{subfigure}
\begin{subfigure}{\columnwidth}
  \centering
  \includegraphics[width=\textwidth]{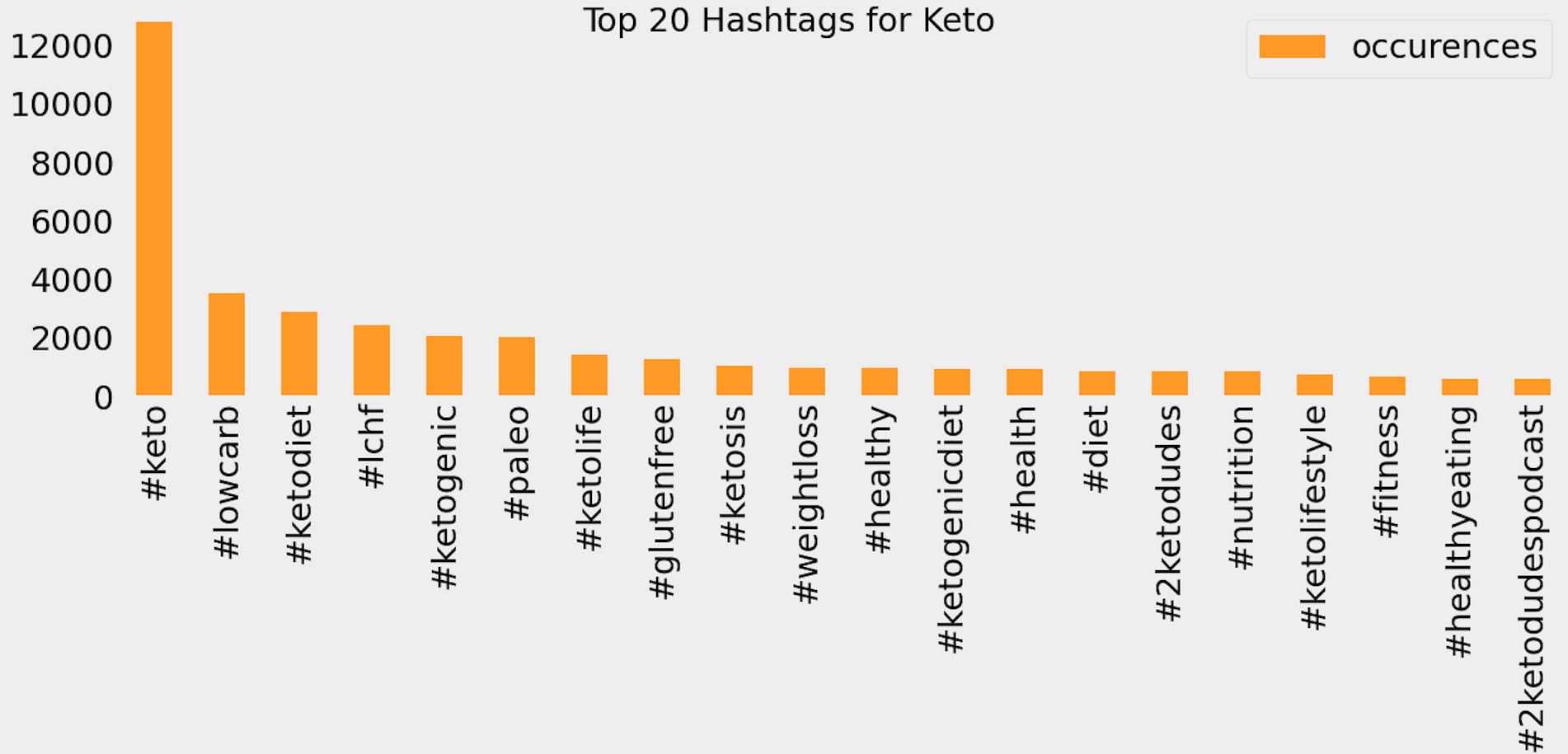}
  \caption{Keto}
  \label{fig:hashtag_keto_bar}
\end{subfigure}
\caption{Top $20$ hashtags related to yoga (green bars) and keto diet (orange bars).}
\label{fig:hashtag}
\end{figure}
To calculate inter-annotator agreement, two graduate students manually annotate a subset of tweets. This subset has an inter-annotator agreement of $64.7\%$ (substantial agreement) using Cohen’s Kappa coefficient \cite{cohen1960coefficient}. 
After checking annotators' disagreement for yoga data, we notice that some yoga teacher/instructor has their yoga studio. They tweet about promoting their yoga studio and sharing their first-hand yoga practice experience. One annotator labels them as practitioners, and another annotates them as promotional. Multi-label classification would be a good option, but we consider only a single label in this paper. Annotators face challenges with user type `others' when they indicate being practitioners (i.e., \textit{yoga lover}) in their Twitter profile description. Still, they retweet about yoga most of the time rather than sharing their experiences in the particular timeline tweets. The annotators then come to a fair agreement to label the remaining tweets of the dataset.
%mention them as practitioners in their Twitter profile description but they retweet about yoga most of the time rather than sharing their experiences in their particular timeline tweets. The annotators then come to a fair agreement to label the remaining tweets of the dataset.

%The annotators then discussed and agreed on general guidelines, which were used to label the remaining tweets of the dataset. 

% For the ground-truth, we manually annotate users (i) for yoga and Keto user type, we label practitioner as `0', promotional as `1', others as `2'; (ii) for yoga user motivation, we annotate health as `0', spirituality as `1', others `2'.

\subsection{Data Distribution} 
In our user type annotated yoga data, we have $42\%$ practitioner, $21\%$ promotional, and $37\%$ other users who love to tweet/retweet about yoga but do not practice yoga. In the user motivation annotated data, we have $51\%$ users who tweet about yoga regarding health benefit, $5\%$ spiritual, and $41\%$ other motivation e.g., business. After annotating $1300$ keto users, we have $50.8\%$ practitioners, $19\%$ promotional, and $30.2\%$ other users.
\section{Experimental Settings}
To run the experiment, we shuffle our dataset and then randomly split it into train (60\%), validation (20\%), and test (20\%).
% \begin{table}
% \begin{adjustbox}{width=\columnwidth,center}
% \Large
% \begin{tabular}{|c|c|c||c|c|}
%           \cline{2-5} \multicolumn{1}{c|}{} & \multicolumn{2}{c||}{\textbf{user type}} & \multicolumn{2}{c|}{\textbf{user motivation}} \\
%           \hline
%     \textbf{Model} & \textbf{Accuracy}  & \textbf{Macro-avg F1}  & \textbf{Accuracy}  & \textbf{Macro-avg F1}  \\
%     \hline
%     Description  & 0.694  & 0.611 & 0.707 & 0.523 \\
%     \hline
%     Location  & 0.639 & 0.520 & 0.694 & 0.517 \\
%     \hline
%     Tweets  & 0.795 & 0.704  & 0.786 & 0.595 \\
%     \hline
%     Network  & 0.726 & 0.561  & 0.798 & 0.590 \\
%     \hline
%     Des\_BF  &  0.718 & 0.681 &  0.771 & 0.528 \\
%     \hline
%     Loc\_BF  &   0.679 & 0.606 & 0.695 & 0.476 \\
%     \hline
%     Twts\_BF  & 0.760 &  0.669  &  0.805 &  0.551 \\
%     \hline
%     Des + Loc & 0.734 & 0.653 & 0.806 & 0.661 \\
%     \hline
%     Des + Net & 0.808 & 0.702 & 0.823 & 0.653 \\
%     \hline
%     Des + Loc + Twt  & 0.778 & 0.705 & 0.808  & 0.603 \\
%     \hline
%     Des + Loc + Net & 0.774 & 0.725 & 0.806  & 0.663 \\
%     \hline
%     % Word2Vec based & 0.790 & 0.742 & 0.844 & 0.61 \\ 
%     % joint embedding &  &  &  & \\
%     Word2Vec based & \multirow{2}{*}{0.790} & \multirow{2}{*}{0.742} & \multirow{2}{*}{0.844} & \multirow{2}{*}{0.61} \\ 
%     joint embedding &  &  &  & \\
    
%     \hline
%     \textbf{Our Model} & \textbf{0.802} & \textbf{0.757} & \textbf{0.853} & \textbf{0.708}  \\
%     \hline
%     \end{tabular}%
% \end{adjustbox}
% \caption{Performance comparisons on yoga data.}
% \label{tab:result}
% \end{table}
\subsection{Baseline Models}
In our experiments, we evaluate our model under twelve different settings: (i) Description; (ii) Location; (iii) Tweets; (iv) Network; (v) BERT fine-tuned with Description (Des\_BF); (vi) BERT fine-tuned with Location (Loc\_BF); (vii) BERT fine-tuned with Tweets (Twts\_BF); (viii) joint embedding on description and location (Des + Loc); (ix) joint embedding on description and network (Des + Net); (x) joint embedding on description, location, and tweets (Des + Loc + Twt); (xi) joint embedding on description, location, and network (Des + Loc + Net), (xii) Word2Vec based joint embedding on description, location, tweets, and network \cite{islam2021you}.
%Description, Location, Tweets, Network, Des\_BF, Loc\_BF, Twts\_BF, Des + Loc, Des + Net, Des + Loc + Twt, Des + Loc + Net, Word2Vec based joint embedding. 

%\footnote[3]{The details description of baseline models is in the supplementary material.}
\subsubsection{Description} 
For user description embedding, we use pre-trained uncased BERTbase model using a masked language modeling (MLM) objective. Transformers in BERT consist of multiple layers, each of which implements a self-attention sub-layer with multiple attention heads. We pass the embedding to stacked Bi-LSTM with a dropout value $0.5$. We get the  hidden representation of description by concatenating the forward and backward directions with dropout ($0.5$). We forward the description representation to one-layer classifier activated by softmax.

\subsubsection{Location} 
To represent location, we follow the same approach as user description. 

\subsubsection{Tweets} 
For user tweets, we concatenate all yoga-related tweets. Then we use pre-trained longformer-base-4096 model for tweet embedding.  We forward the tweet embedding to stacked Bi-LSTM with dropout layer ($0.5$). We get the hidden representation of tweets by concatenating the forward and backward directions with dropout ($0.5$). To assign important words in the final representation, we use a context-aware attention mechanism. We forward the tweet representation to one-layer classifier activated by softmax.

\subsubsection{Network} 
We use Node2Vec for network embedding after the construction of the user network. We forward the network embedding to a linear layer for obtaining network representation. We pass the network representation to a dropout layer of value $0.5$ with ReLU activation and then forward to one-layer classifier activated by softmax.

\subsubsection{Des + Loc} 
We concatenate profile description and location representations and pass through a two-layer classifier activated by ReLU and softmax, respectively.

\subsubsection{Des + Net} 
We concatenate user's profile description and user network representations and forward the joint representation to a two-layer classifier activated by ReLU and then softmax.

\begin{table*}
\centering
%\captionsetup{font=Large}
%\begin{adjustbox}{width=1.75\columnwidth,center}
%\Large
\begin{tabular}{lcccc}
    \toprule
    \multirow{2}{*}{\textbf{Model}} & \multicolumn{2}{c}{\textbf{User type}} & \multicolumn{2}{c}{\textbf{User motivation}} \\
    \cmidrule(r){2-3}\cmidrule(l){4-5}
     & \textbf{Accuracy}  & \textbf{Macro-avg F1}  & \textbf{Accuracy}  & \textbf{Macro-avg F1}  \\
     \midrule
    Description  & 0.694  & 0.611 & 0.707 & 0.523 \\
    Location  & 0.639 & 0.520 & 0.694 & 0.517 \\
    Tweets  & 0.795 & 0.704  & 0.786 & 0.595 \\
    Network  & 0.726 & 0.561  & 0.798 & 0.590 \\
    Des\_BF  &  0.718 & 0.681 &  0.771 & 0.528 \\
    Loc\_BF  &   0.679 & 0.606 & 0.695 & 0.476 \\
    Twts\_BF  & 0.760 &  0.669  &  0.805 &  0.551 \\
    Des + Loc & 0.734 & 0.653 & 0.806 & 0.661 \\
    Des + Net & 0.808 & 0.702 & 0.823 & 0.653 \\
    Des + Loc + Twt  & 0.778 & 0.705 & 0.808  & 0.603 \\
    Des + Loc + Net & 0.774 & 0.725 & 0.806  & 0.663 \\
    % Word2Vec based & 0.790 & 0.742 & 0.844 & 0.61 \\ 
    % joint embedding &  &  &  & \\
    Word2Vec based joint embedding & 0.790 & 0.742 & 0.844 & 0.610 \\[.2em]
    \textbf{Our Model} & \textbf{0.802} & \textbf{0.757} & \textbf{0.853} & \textbf{0.708}  \\
    \bottomrule
    \end{tabular}%
%\end{adjustbox}
\caption{Performance comparisons on yoga data.}
\label{tab:result}
\end{table*}

\subsubsection{Des + Loc + Twt} 
We concatenate description, location, tweet representations and pass the joint representation to a two-layer classifier activated by ReLU and softmax correspondingly.

\subsubsection{Des + Loc + Net} 
In this case, concatenation of description, location, and user network representations are fed to the two-layer classifier with ReLU and softmax activation function.

\subsubsection{Fine-tuning pre-trained BERT}
BERT uses bidirectional transformers to pre-train a large corpus and fine-tunes the pre-trained model on other tasks. We use description, location, and tweets separately and fine-tune the pre-trained BERT (base-uncased) for Des\_BF, Loc\_BF, and Twts\_BF baselines. 
% For BERT fine-tuning, \textit{BertForSequenceClassification} model with an added single linear layer with softmax activation on top is used for the classification task of user type and user motivation.

%\subsubsection{Des\_BF}
\textbf{Des\_BF}
For this setting, we use a pre-trained BERT (\textit{BertForSequenceClassification}) model and fine-tune it with an added single linear layer on top. In this case, BERT's input is the user's profile description constructed by the summation of the corresponding token, segment, and position embeddings for a given token.  

%\textit{BertForSequenceClassification} model with an added single linear layer on top is used for the classification task of user type and user motivation. As we feed input data, the entire pre-trained BERT model and the additional untrained classification layer is trained on our specific task. 

% We use the final hidden state of the first token as the input. We denote the vector as $F \in \mathbb{R}^h$ . Then we add a classifier whose parameter matrix is $W \in \mathbb{R}^{l \times h}$, where $l$ is the number of class label. Finally, the probability $P$ of each class label is calculated by the softmax function $P = softmax(FW^T)$.

%\subsubsection{Loc\_BF}
\textbf{Loc\_BF}
We use a pre-trained and fine-tuned BERT model with user location to classify user type and user motivation. 
 
%\subsubsection{Twts\_BF}
\textbf{Twts\_BF}
For Twts\_BF model, we use a pre-trained and fine-tuned \textit{BertForSequenceClassification} model with user tweets to classify our tasks. 

\subsubsection{Word2Vec based joint embedding}
Instead of using pre-trained BERT, we use pre-trained Word2Vec \cite{mikolov2013efficient} for tweets, location, and description embedding. Then concatenate the four sub-networks description, location, tweets, and user networks and pass them to the two-layer classifier with ReLU and softmax activation function.

\subsection{Hyperparameter Details}
We set the hyperparameters of our final model as follows:
batch size = $32$, learning rate = $0.001$, epochs = $10$. The dropout rate between layers is set to $0.5$. 
%For all the models except the BERT fine-tuned models, 
We perform grid hyperparameter search on the validation set using early stopping for all the models except these three models -- Des\_BF, Loc\_BF, Twts\_BF. For learning rate, we investigate values $0.001, 0.01, 0.05, 0.1 $; for $L^2$ regularization, we examine $0, 10^{-3}, 10^{-2}$ values; and for dropout, values $0.2, 0.25, 0.4, 0.5$. We run the models total $10$ epochs and plot curves for loss, accuracy, and macro-avg F1 score. Our early stopping criterion is based on the validation loss when it starts to increase sharply\footnote[2]{The plots are in the supplementary material.}. For Word2Vec based joint embedding model, we have the same hyperparameters used by \cite{islam2021you}.

%For network embeddings as well as Des\_BF, Loc\_BF, and Twts\_BF model, we have the same parameters used by \cite{islam2021you}.
%A summary of hyperparameter settings of the models is shown in Table \ref{tab:hyperparam}. 

In the Des\_BF, Loc\_BF, and Twts\_BF models, for padding or truncating text, we chose maximum sentence length = $160, 50, 500$ correspondingly. We use learning rate = $2e-5$, optimizer= AdamW \cite{loshchilov2018decoupled}, epochs = $4$, epsilon parameter = $1e-8$, batch size = $32$,

Network embeddings are trained using Node2Vec with following parameters: dimension = $300$, number of walks per source = $10$, length of walk per source = $80$, minimum count = $1$, window size = $10$ and then forwarded to a linear layer of size $150$. If users do not appear in the $@$-mentioned network, we set their network embedding vectors $0$. Also, for the users without having location and/or description, we set the embedding vectors as $0$ correspondingly.
%\footnote[5]{Hyperparameter details in the supplementary material.} 
Table \ref{tab:hyperparam} summarizes the hyperparameter settings of all models. 

% Fig. \ref{fig:byun_loss_f1} shows the learning curves loss (train and validation) vs. epochs, macro-avg F1 score (train and validation) vs. epochs, and accuracy (train and validation) vs. epochs for BYUN model.

\section{Results and Analysis}
%\label{subsec:5.5}
As reported in Table \ref{tab:result}, our proposed model achieves the best test accuracy and macro-avg F1 score for classifying yoga user type and motivation and outperforms the baseline models. Our model obtains the highest test accuracy $(80.2\%)$ and macro-avg F1 score $(75.7\%)$ for classifying yoga user type (Table \ref{tab:result}). We achieve noticeable performance for classifying yoga user motivation where our model obtains the highest test accuracy $(85.3\%)$ and macro-avg F1 score $(70.8\%)$. 
%We have $14.4\%$ improvement in macro-avg F1 score of predicting user motivation. 
To predict user type in keto data, we get test accuracy $(71.9\%)$ and macro-avg F1 score $(67.6\%)$.

%Fig. \ref{fig:yun_loss_f1} shows the learning curves loss (train and validation) vs. epochs and macro-avg F1 score (train and validation) vs. epochs for BYUN model.
% \vspace{-.7pt}
\begin{figure}[t]
\begin{subfigure}{.5\columnwidth}
  \centering
  \includegraphics[width=\textwidth]{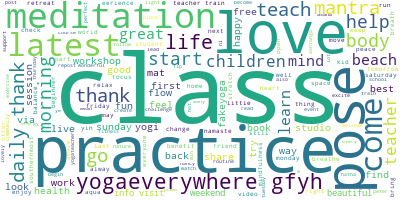}
  \caption{}
  \label{fig:prac_yoga_wo}
\end{subfigure}%
\begin{subfigure}{.5\columnwidth}
  \centering
  \includegraphics[width=\textwidth]{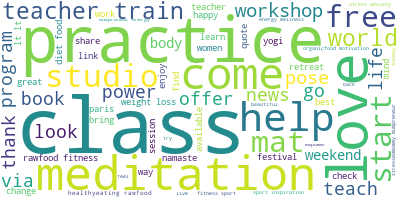}
  \caption{}
  \label{fig:promo_yoga_wo}
\end{subfigure}
\begin{subfigure}{.5\columnwidth}
  \centering
  \includegraphics[width=\textwidth]{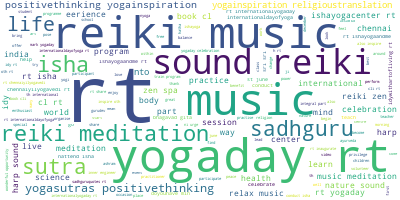}
  \caption{}
  \label{fig:other_yoga_wo}
\end{subfigure}%
\begin{subfigure}{.5\columnwidth}
  \centering
  \includegraphics[width=\textwidth]{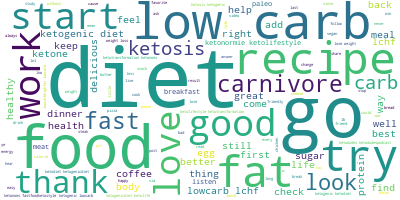}
  \caption{}
  \label{fig:prac_keto_wo}
\end{subfigure}
\begin{subfigure}{.5\columnwidth}
  \centering
  \includegraphics[width=\textwidth]{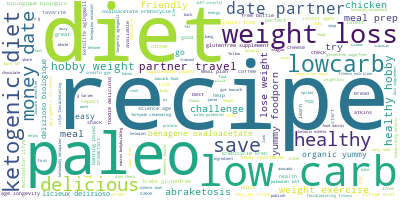}
  \caption{}
  \label{fig:promo_keto_wo}
\end{subfigure}%
\begin{subfigure}{.5\columnwidth}
  \centering
  \includegraphics[width=\textwidth]{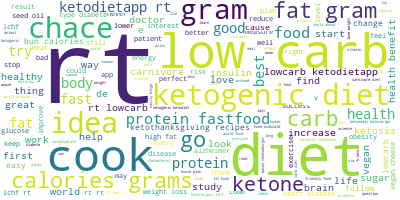}
  \caption{}
  \label{fig:other_keto_wo}
\end{subfigure}
\caption{Wordcloud for yoga and keto users' tweets. (a) yoga: practitioner, (b) yoga: promotional, (c) yoga: others, (d) keto: practitioner, (e) keto: promotional, (f) keto: others.}
\label{fig:yoga_keto_wc}
\end{figure}
\subsection{Ablation Study}
We train an individual neural network model for each field -- description, location, tweets, and network. The $1^{st}$ four rows of the Table \ref{tab:result} show the performance breakdown for each model over the test dataset. The results conclude that tweets and profile description of any user is informative fields for our task. 
However, our experiments show that either excluding user network information (Des + Loc + Twt model) or tweets information (Des + Loc + Net model) declines the final model's performance in terms of both accuracy and macro-avg F1 for both user type and motivation classification task ($10^{th}$ and $11^{th}$ rows of Table \ref{tab:result}). 

\begin{figure}[t]
\begin{subfigure}{.5\columnwidth}
  \centering
  \includegraphics[width=\textwidth]{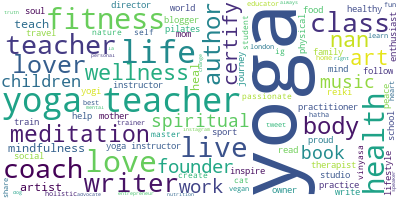}
  \caption{}
  \label{fig:prac_yoga_desc}
\end{subfigure}%
\begin{subfigure}{.5\columnwidth}
  \centering
  \includegraphics[width=\textwidth]{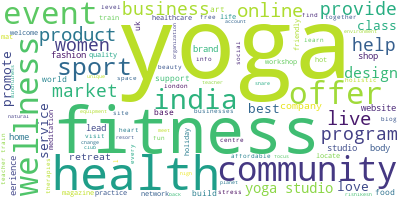}
  \caption{}
  \label{fig:promo_yoga_desc}
\end{subfigure}
\begin{subfigure}{.5\columnwidth}
  \centering
  \includegraphics[width=\textwidth]{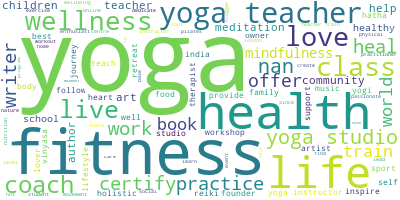}
  \caption{}
  \label{fig:health_yoga_desc}
\end{subfigure}%
\begin{subfigure}{.5\columnwidth}
  \centering
  \includegraphics[width=\textwidth]{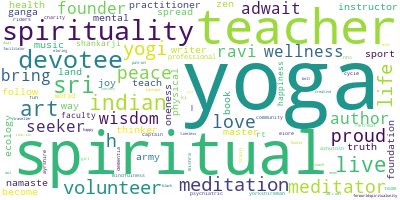}
  \caption{}
  \label{fig:spirit_yoga_desc}
\end{subfigure}
\begin{subfigure}{.5\columnwidth}
  \centering
  \includegraphics[width=\textwidth]{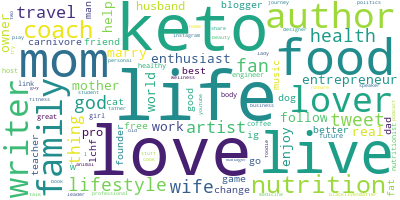}
  \caption{}
  \label{fig:prac_keto_desc}
\end{subfigure}%
\begin{subfigure}{.5\columnwidth}
  \centering
  \includegraphics[width=\textwidth]{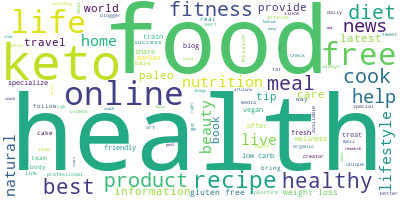}
  \caption{}
  \label{fig:promo_keto_desc}
\end{subfigure}
\caption{Wordcloud for yoga and keto users' profile description. (a) yoga: practitioner, (b) yoga: promotional, (c) yoga: health motivation, (d) yoga: spiritual motivation, (e) keto: practitioner, (f) keto: promotional.}
\label{fig:yoga_keto_desc_wc}
\end{figure}
\subsection{Top Hashtags}
Fig. \ref{fig:hashtag} shows the top $20$ hashtags $(\#)$ related to yoga and keto diet and the number of occurrences of those hashtags in our data. Most of them are self-explanatory. In the yoga dataset, the popular hashtag $\#namaste$ is used by the users meaning `bow me you' or `I bow to you', some users use $\#gfyh$ representing `Go 4 Yoga Health', hashtag $\#mantra$ translates to `vehicle of the mind'. However, the keto diet is related to a low carb high-fat diet; that's why the common hashtag $\#lchf$. 

\begin{figure*}
\centering
\begin{subfigure}{.91\columnwidth}
  \centering
  \includegraphics[width=\textwidth]{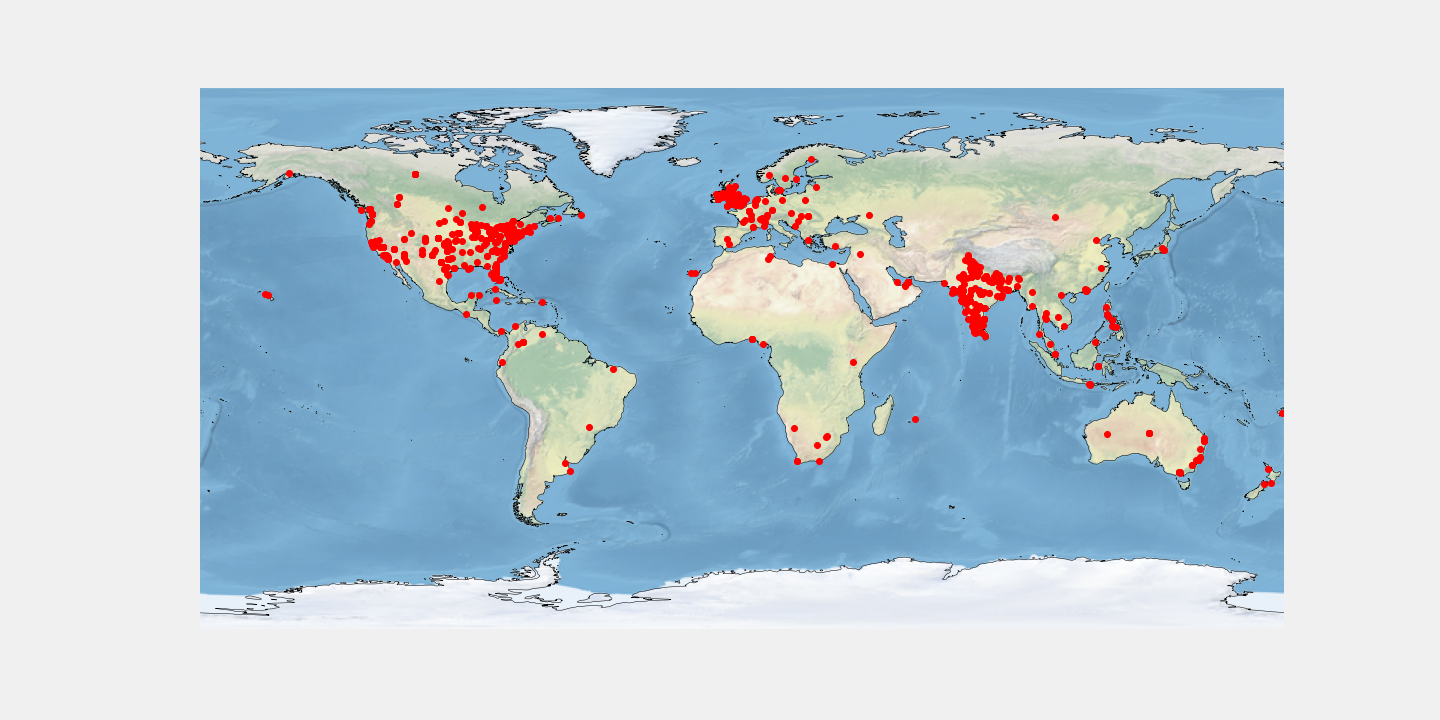}
  \caption{Whole yoga data.}
  \label{fig:all_yoga_map}
\end{subfigure}%
\begin{subfigure}{.91\columnwidth}
  \centering
  \includegraphics[width=\textwidth]{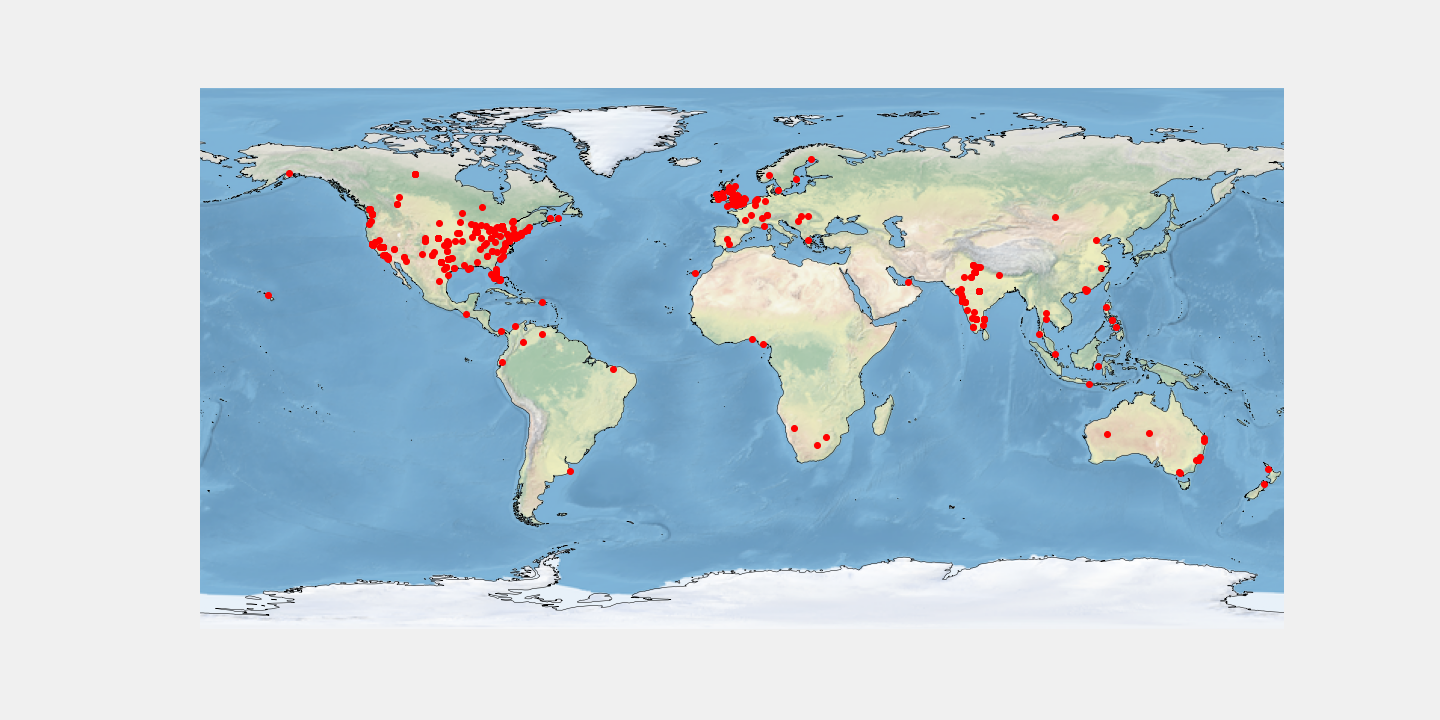}
  \caption{Yoga: practitioner.}
  \label{fig:prac_yoga_map}
\end{subfigure}
\begin{subfigure}{.91\columnwidth}
  \centering
  \includegraphics[width=\textwidth]{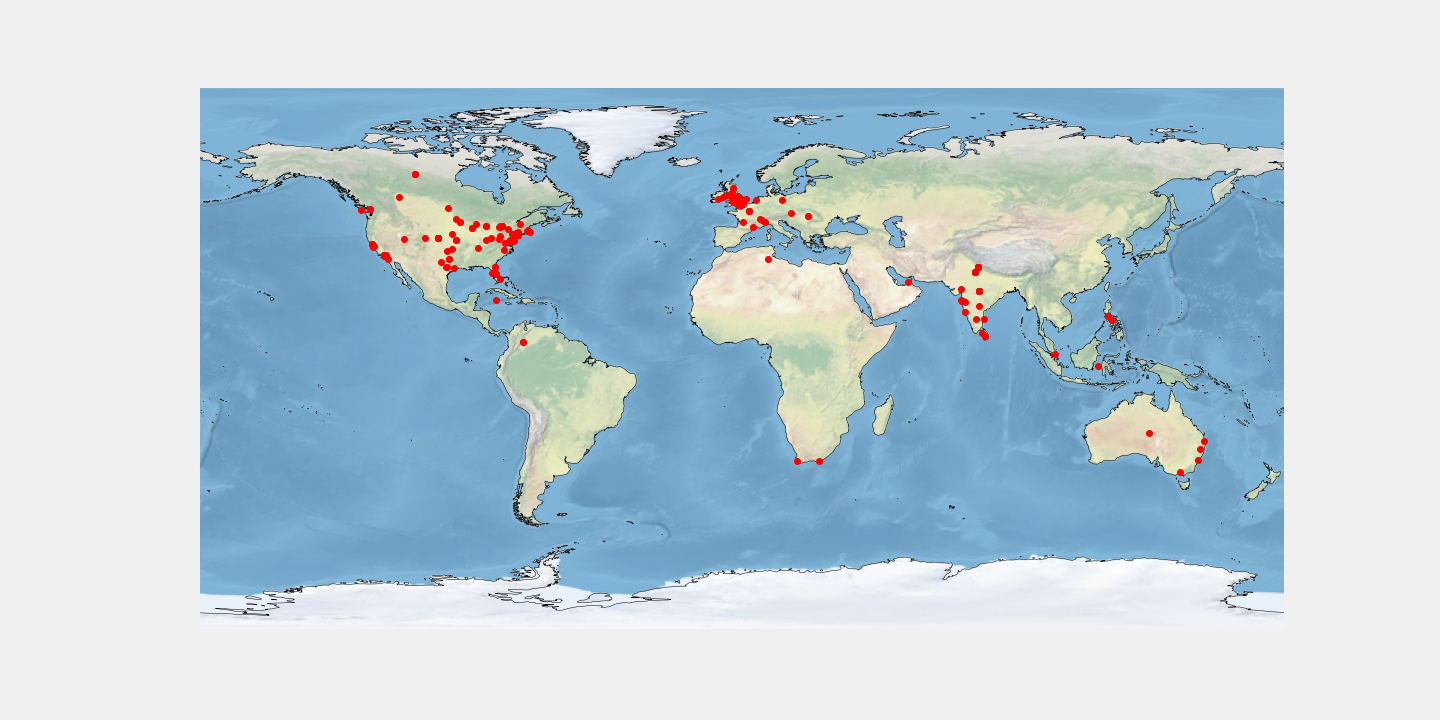}
  \caption{Yoga: promotional.}
  \label{fig:promo_yoga_map}
\end{subfigure}%
\begin{subfigure}{.91\columnwidth}
  \centering
  \includegraphics[width=\textwidth]{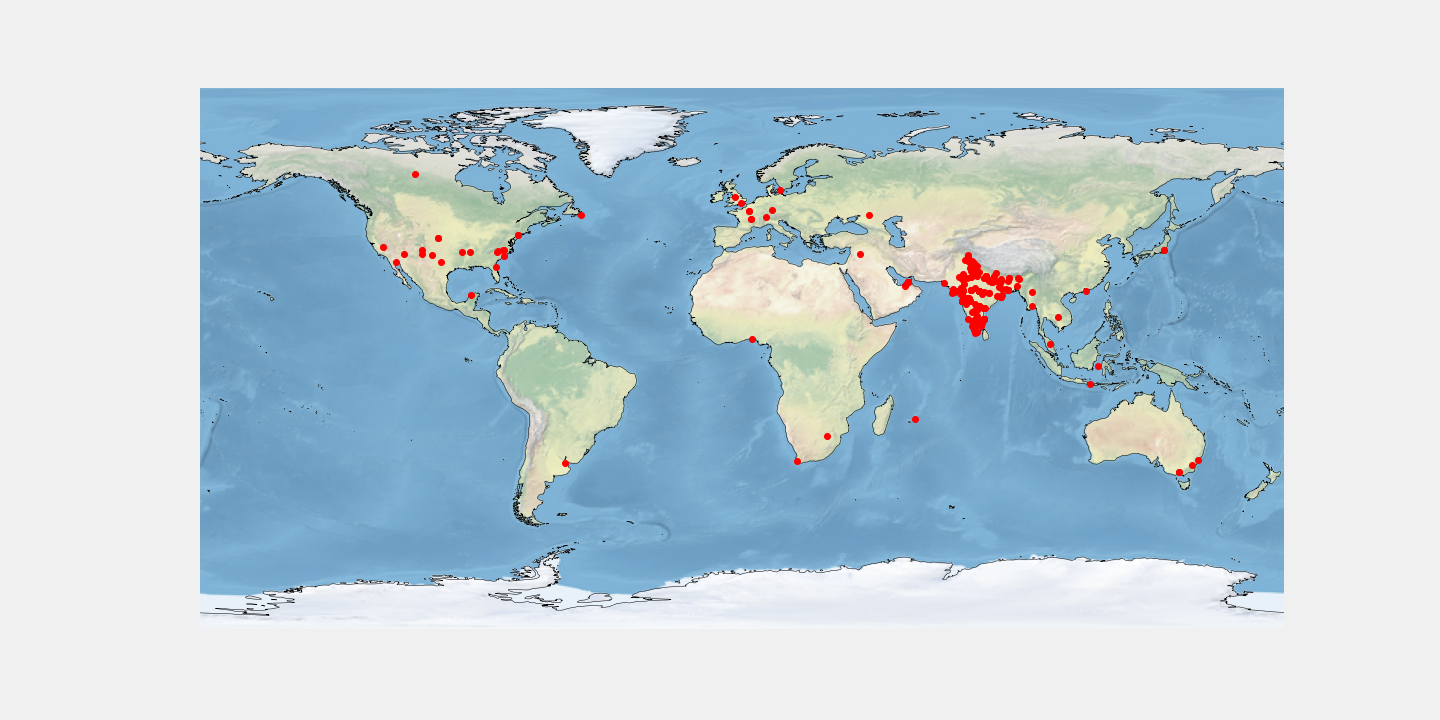}
  \caption{Yoga: others.}
  \label{fig:other_yoga_map}
\end{subfigure}
\begin{subfigure}{.91\columnwidth}
  \centering
  \includegraphics[width=\textwidth]{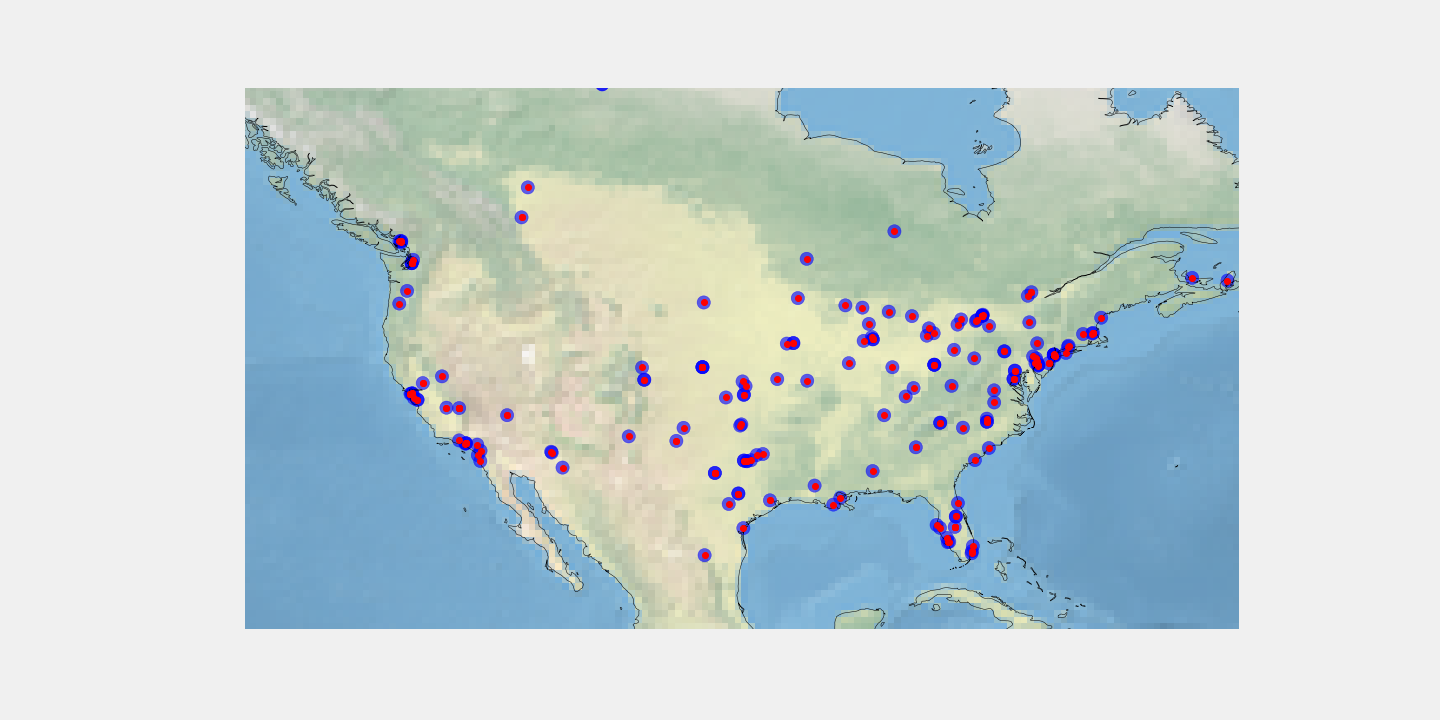}
  \caption{Yoga practitioner from USA.}
  \label{fig:prac_yoga_usa_map}
\end{subfigure}%
\begin{subfigure}{.91\columnwidth}
  \centering
  \includegraphics[width=\textwidth]{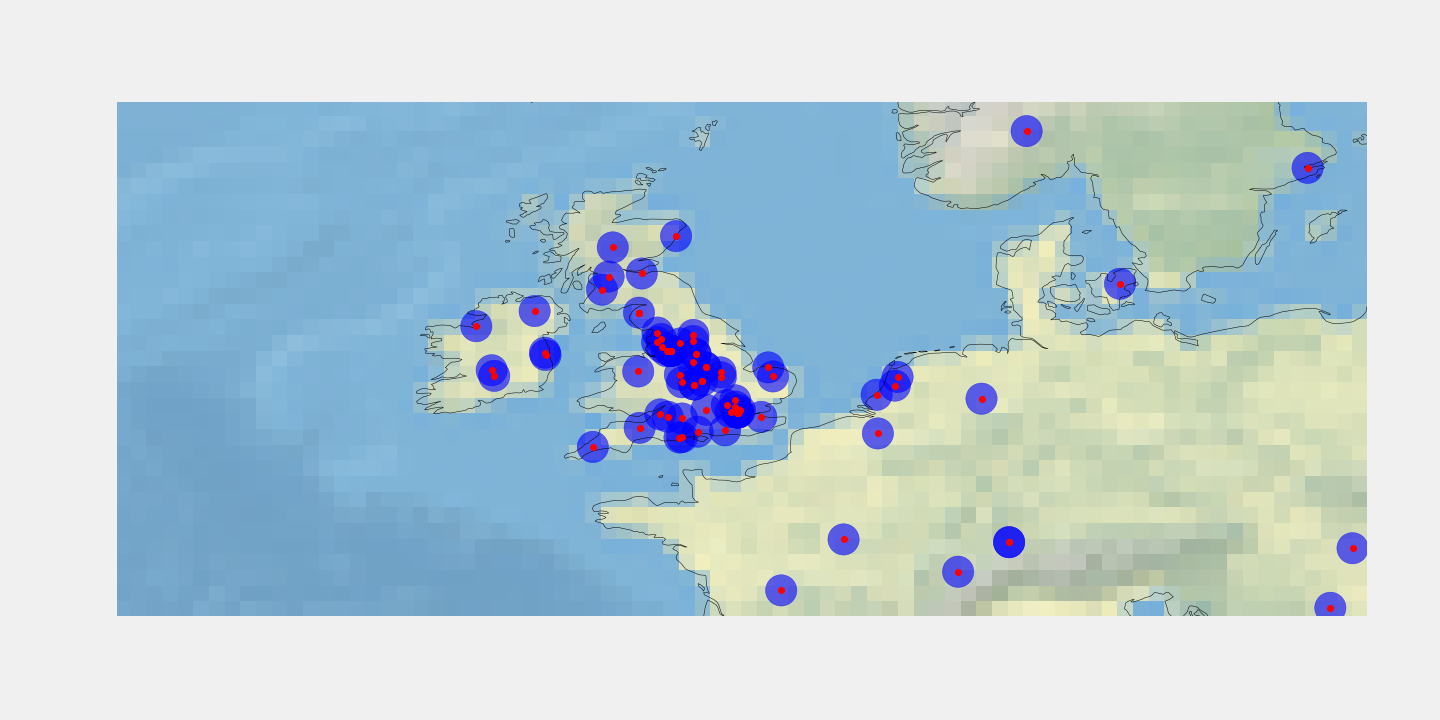}
  \caption{Yoga practitioner from UK.}
  \label{fig:prac_yoga_UK_map}
\end{subfigure}
\begin{subfigure}{.91\columnwidth}
  \centering
  \includegraphics[width=\textwidth]{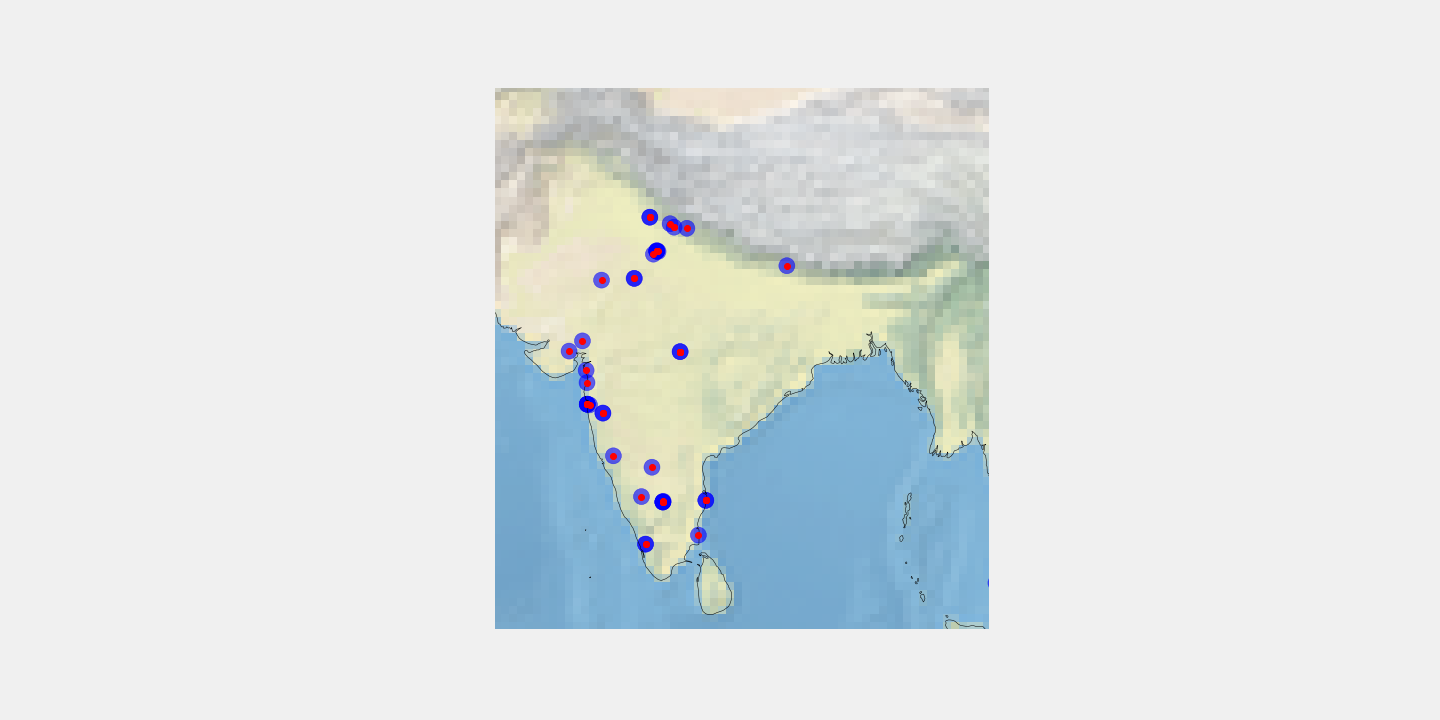}
  \caption{Yoga practitioner from India.}
  \label{fig:prac_yoga_india_map}
\end{subfigure}%
\begin{subfigure}{.91\columnwidth}
  \centering
  \includegraphics[width=\textwidth]{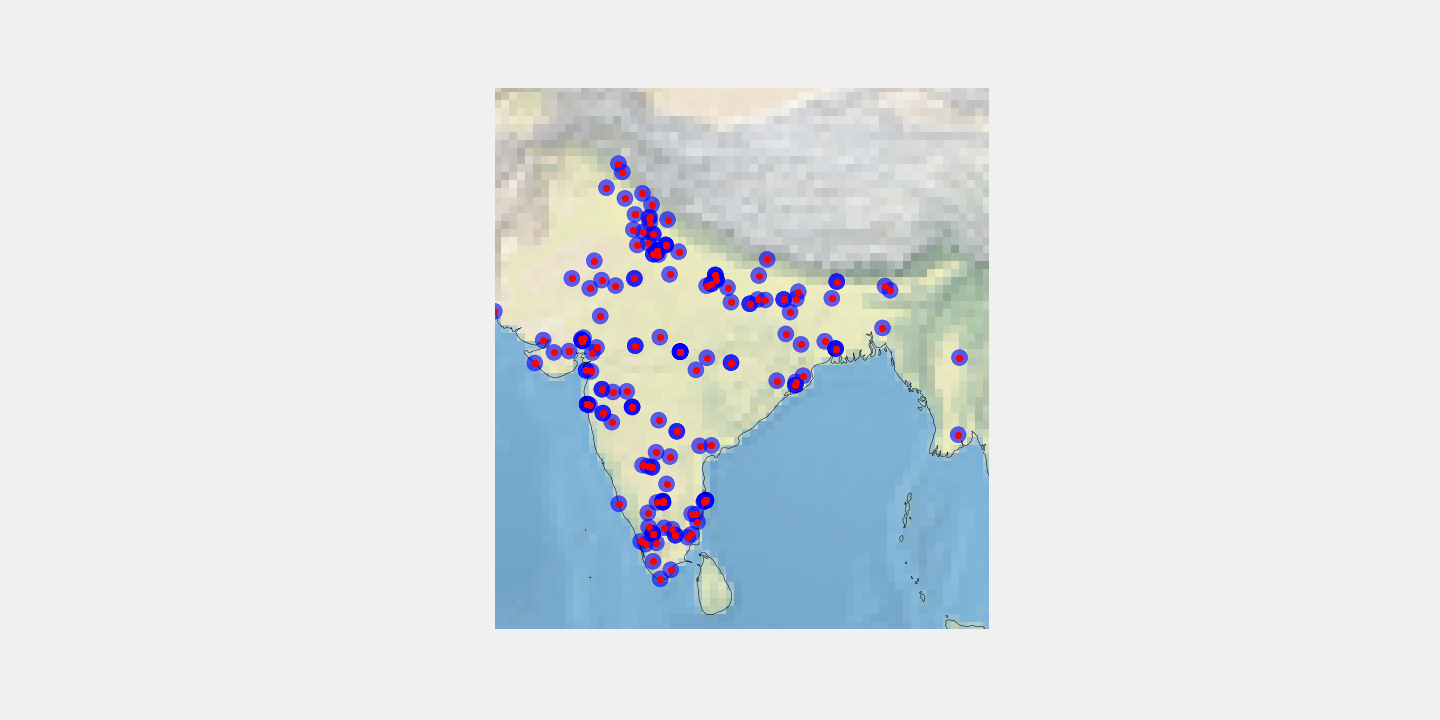}
  \caption{Others from India.}
  \label{fig:other_yoga_india_map}
\end{subfigure}
\begin{subfigure}{.91\columnwidth}
  \centering
  \includegraphics[width=\textwidth]{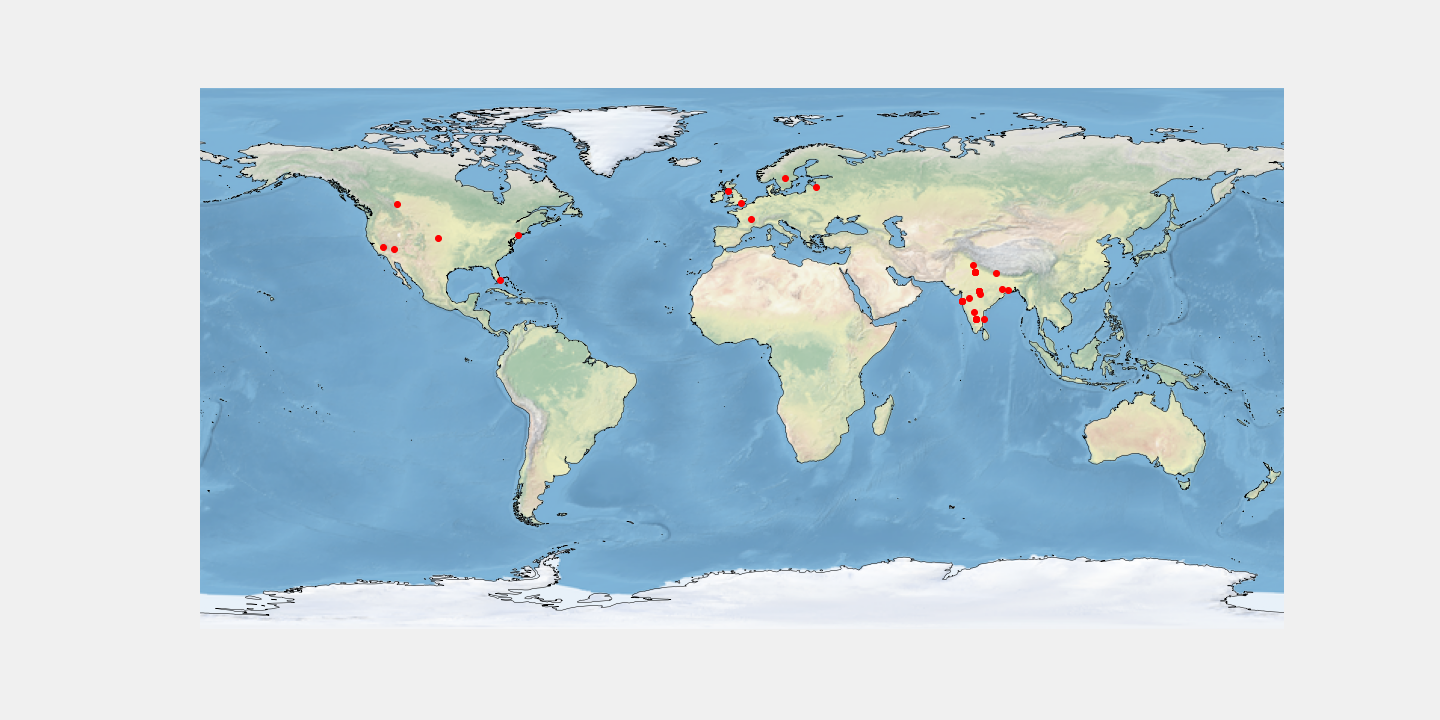}
  \caption{Yoga: spiritual motivation.}
  \label{fig:spirit_yoga_map}
\end{subfigure}%
\begin{subfigure}{.91\columnwidth}
  \centering
  \includegraphics[width=\textwidth]{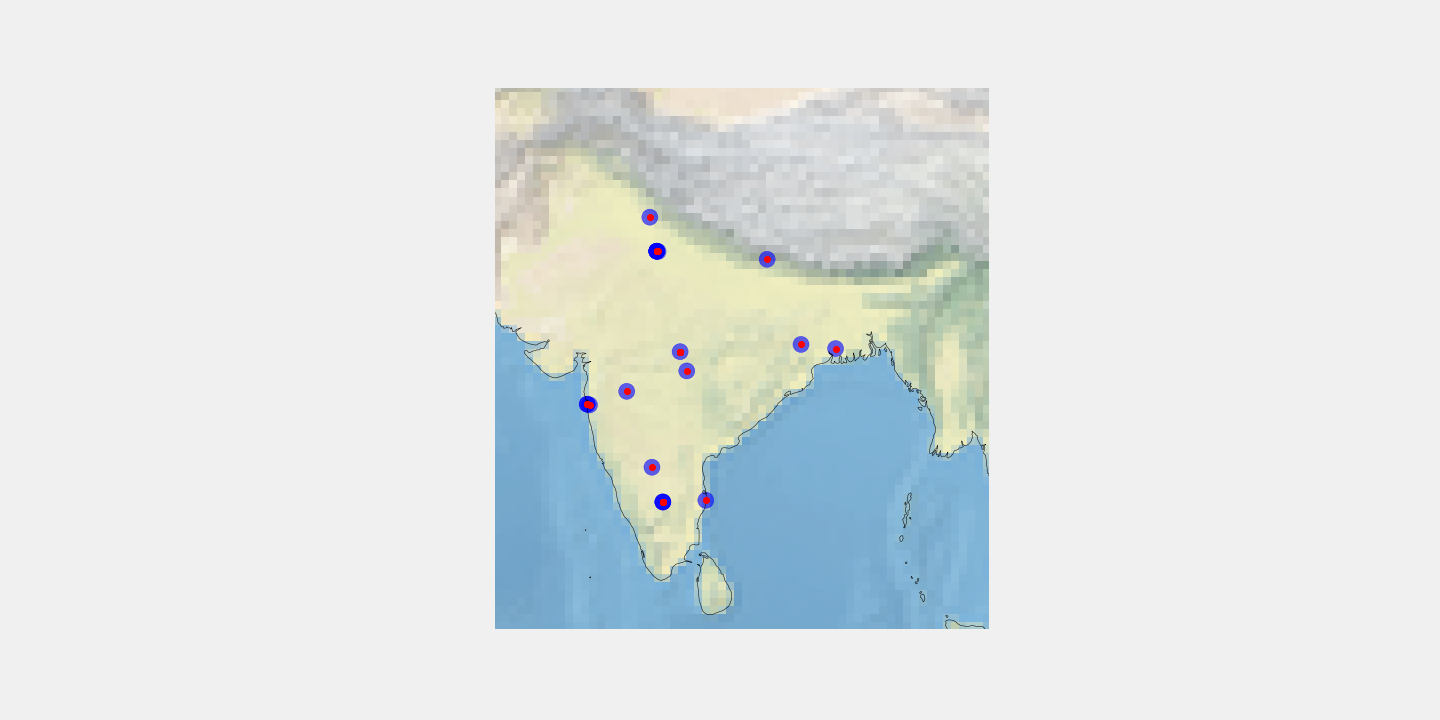}
  \caption{Yoga: spiritual motivation from India.}
  \label{fig:spirit_yoga_india_map}
\end{subfigure}
\caption{Yoga user distribution over location.}
%Yoga user distribution over location. (a) whole yoga data, (b) yoga: practitioner, (c) yoga: promotional, (d) yoga: others, (e) yoga practitioner from USA, (f) yoga practitioner from UK, (g) yoga practitioner from India, (h) others from India, (i) yoga: spiritual motivation, (j) yoga: spiritual motivation from India.}
\label{fig:yoga_map}
\end{figure*}

\begin{figure}
\begin{subfigure}{.5\columnwidth}
  \centering
  \includegraphics[width=\textwidth]{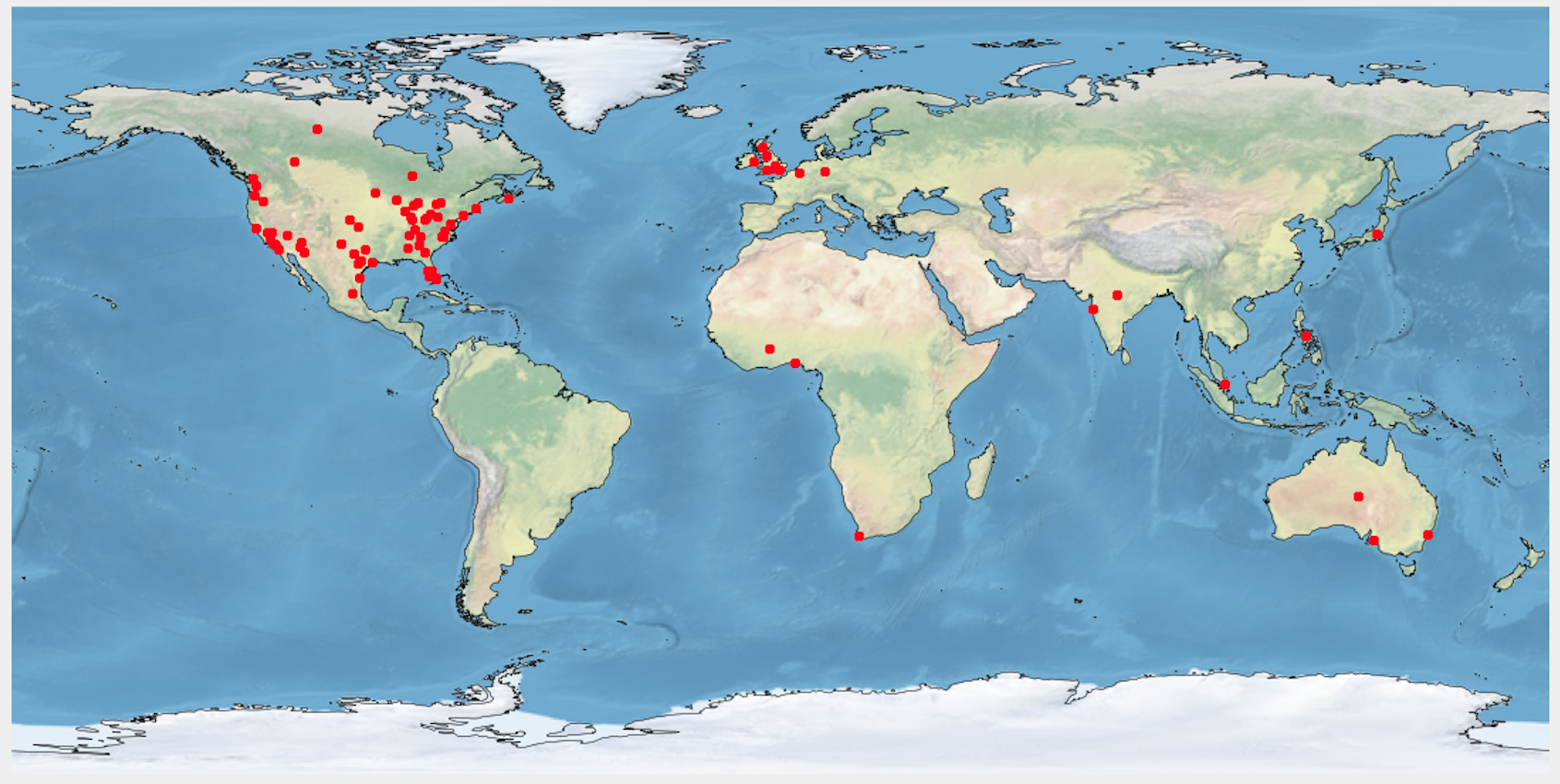}
  \caption{}
  \label{fig:all_keto_map}
\end{subfigure}%
\begin{subfigure}{.5\columnwidth}
  \centering
  \includegraphics[width=\textwidth]{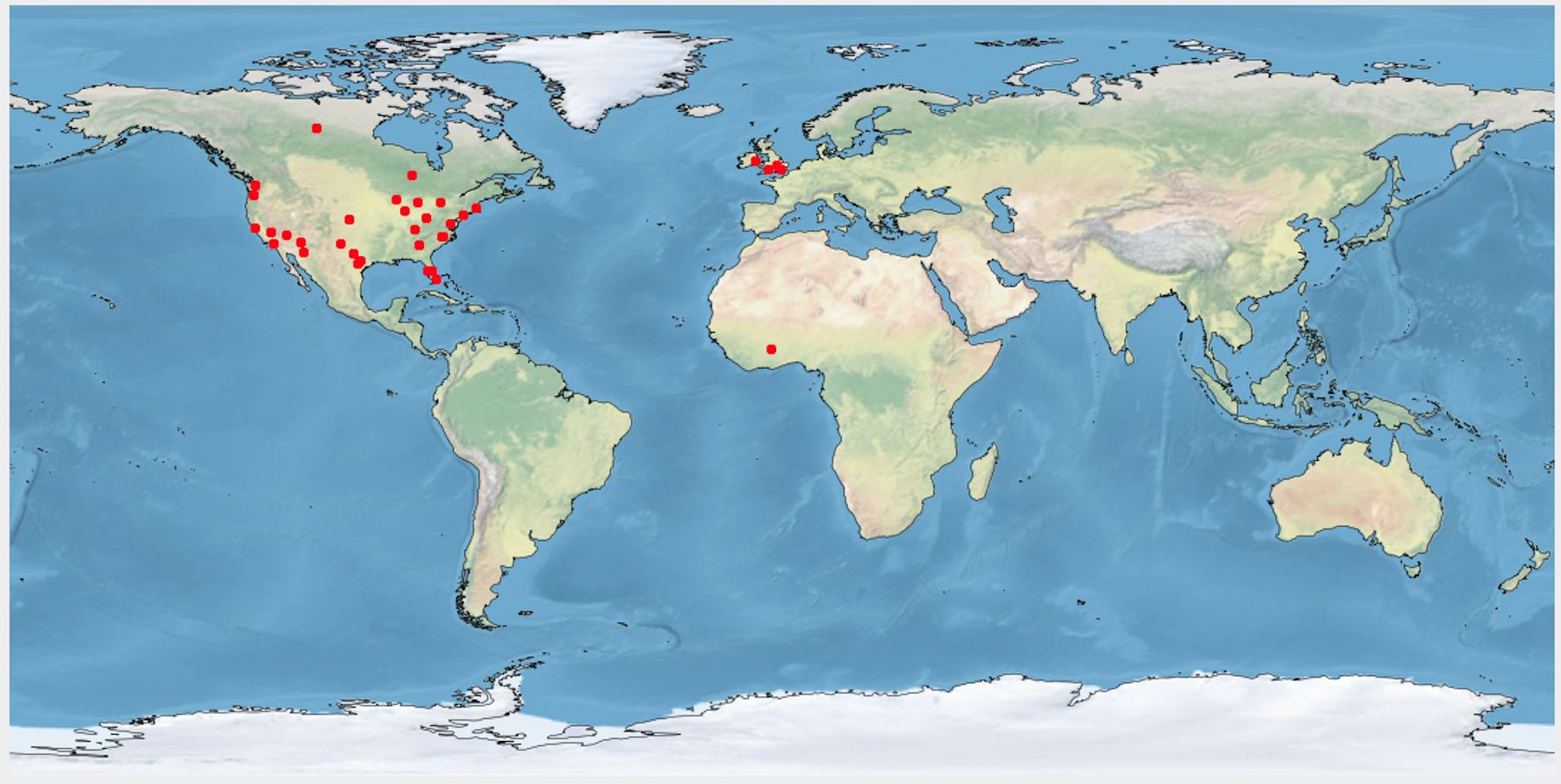}
  \caption{}
  \label{fig:prac_keto_map}
\end{subfigure}
\begin{subfigure}{.5\columnwidth}
  \centering
  \includegraphics[width=\textwidth]{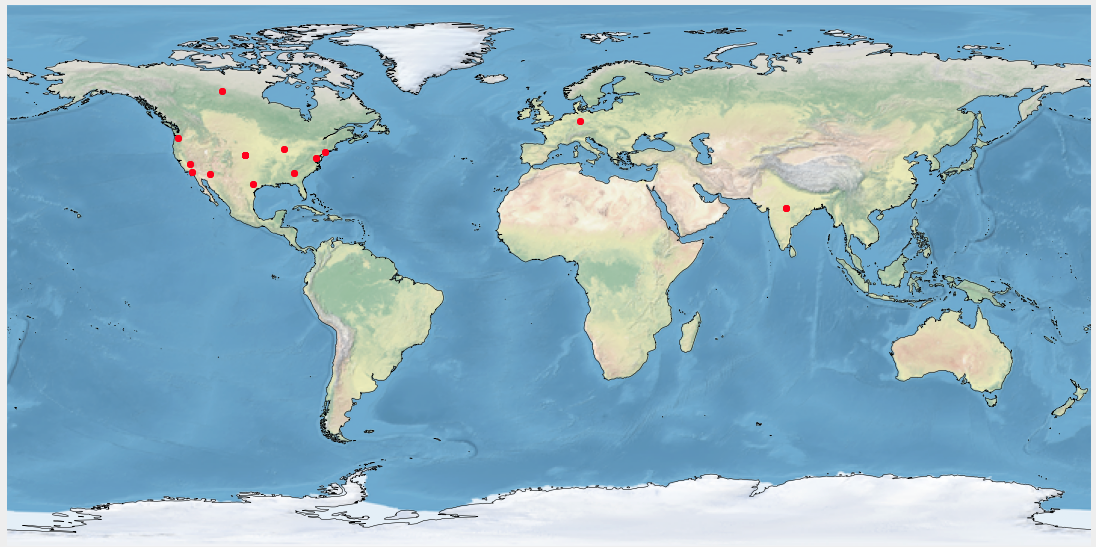}
  \caption{}
  \label{fig:promo_keto_map}
\end{subfigure}%
\begin{subfigure}{.5\columnwidth}
  \centering
  \includegraphics[width=\textwidth]{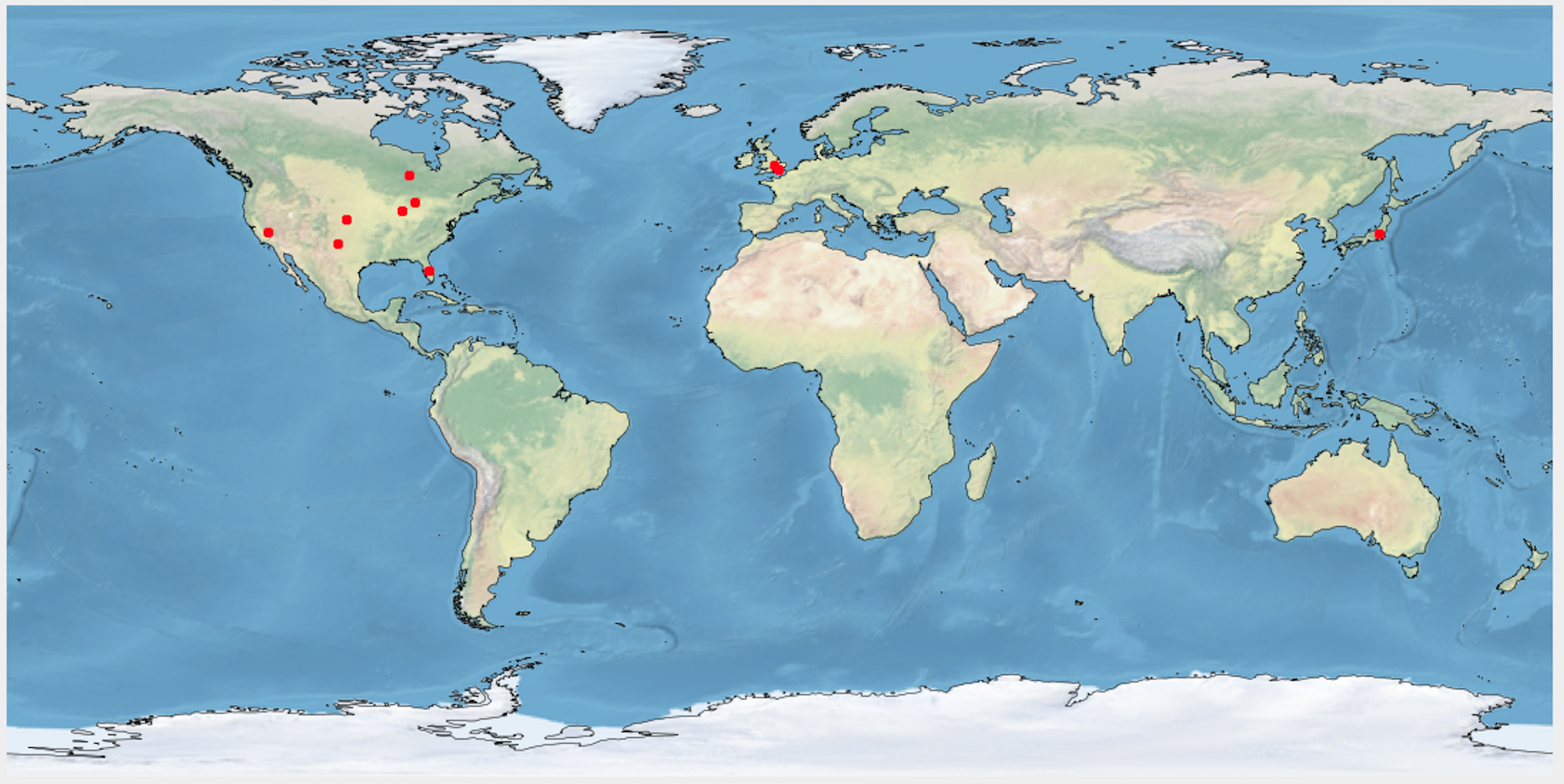}
  \caption{}
  \label{fig:other_keto_map}
\end{subfigure}
\begin{subfigure}{.5\columnwidth}
  \centering
  \includegraphics[width=\textwidth]{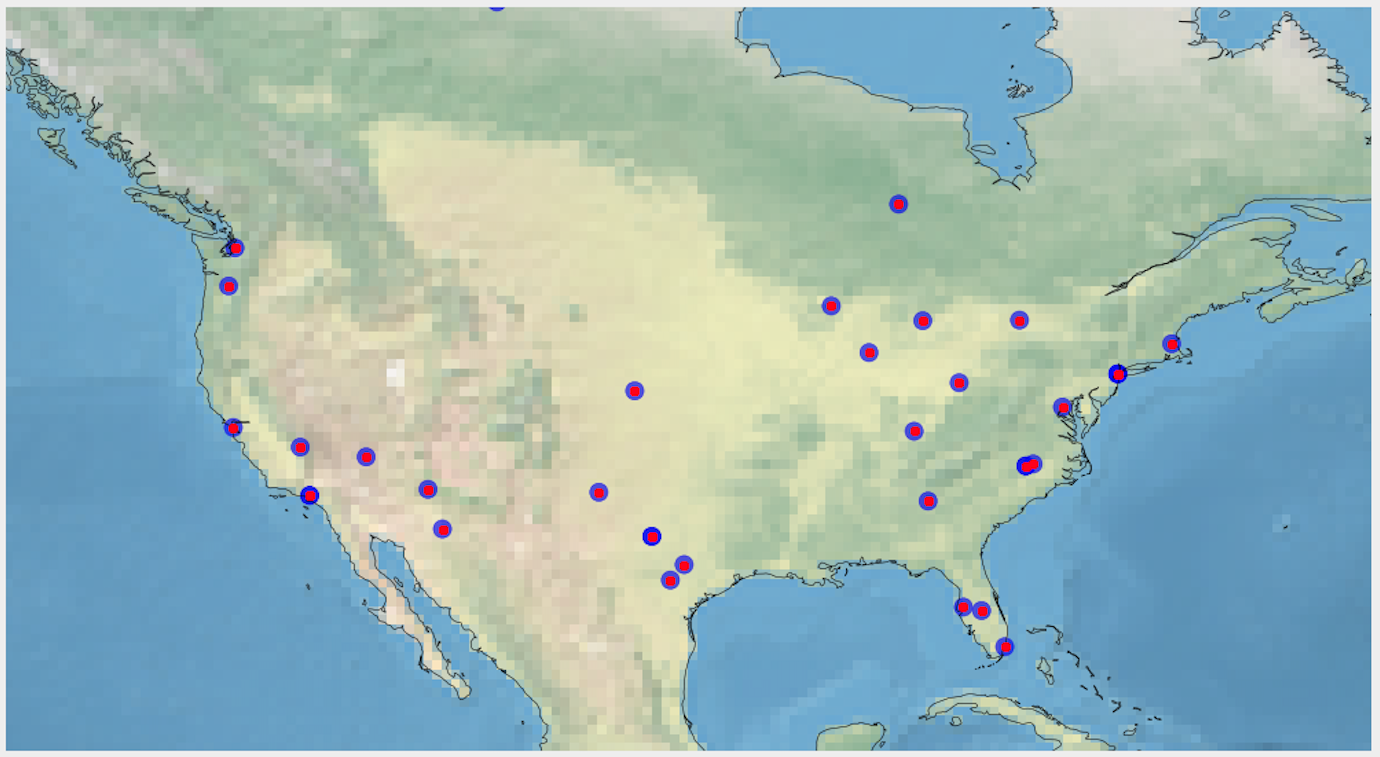}
  \caption{}
  \label{fig:prac_keto_usa_map}
\end{subfigure}%
\begin{subfigure}{.5\columnwidth}
  \centering
  \includegraphics[width=\textwidth]{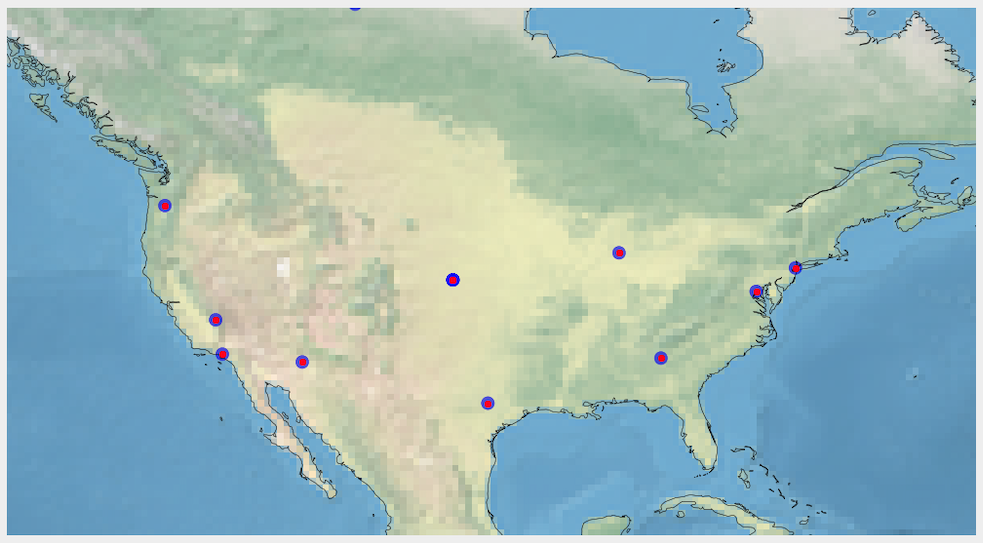}
  \caption{}
  \label{fig:promo_keto_usa_map}
\end{subfigure}
\caption{Keto user distribution over location. (a) whole keto data, (b) practitioner, (c) promotional, (d) others, (e) keto practitioner from USA, (f) keto promotional from USA.}
\label{fig:keto_map}
\end{figure}
\subsection{Relationship between Tweets and Labels}
To understand what kind of words users use in their tweets, we create wordcloud with the most frequent words (Fig. \ref{fig:yoga_keto_wc}) from the yoga and keto dataset. To generate the wordcloud, we filter out the word `yoga' and `keto' from the yoga and keto dataset tweets, respectively, because of the apparent high occurrences. 

We notice that the most frequent words from tweets of yoga practitioners are \textit{practice, love, pose, class, meditation, mind, mantra, daily, thank, yogaeverywhere, gfyh} (Fig. \ref{fig:prac_yoga_wo}). Because some practitioners tweet about `daily yoga practice/class/pose', some of them share `love/thankfulness about yoga', some practitioners tweet with popular \textit{hashtag} i.e., $\#yogaeverywhere$,  $\#gfyh$.

Promotional users have the following words \textit{class, studio, practice, come, train, teacher, workshop, free, mat, offer} (Fig. \ref{fig:promo_yoga_wo}). In most cases, promotional yoga users i.e., studio/gym tweets about `offering to teach/train free yoga class/workshop', online shops tweet about `selling yoga mat'.

Other users mostly retweet and share news of yoga/yogi rather than directly practicing or promoting yoga. They have noticeable words such as \textit{rt, reiki, sadhguru, isha, yogaday} (Fig. \ref{fig:other_yoga_wo}) where \textit{rt} stands for retweet, \textit{reiki} is a soothing yoga treatment, \textit{isha foundation} is a non-profit organization in India by \textit{Sadhguru (yogi) Jaggi Vasudev}. As most of the `others' user in our data are from India (Fig. \ref{fig:other_yoga_map} and \ref{fig:other_yoga_india_map}), the reason for those words in the worldcloud is understandable.

For keto practitioner's tweets, we observe that most frequent words are \textit{diet, low carb, fat, carnivore, ketosis, food, recipe, start, go, try, love, thank, fast, protein, meat, egg} (Fig. \ref{fig:prac_keto_wo}). As some practitioners tweet about `starting of their keto lifestyle', some of them advise others to `try/go for keto', some practitioners tweet about `keto recipe'. Another popular term called `keto carnivore' takes the ketogenic diet to an animal protein-based diet, i.e., `meat', `egg'.

Tweets from promotional keto users have the following words \textit{recipe, diet, paleo, low carb, weight loss, delicious, meal prep, money, healthy, organic, yummy} (Fig. \ref{fig:promo_keto_wo}). In most cases, promotional keto users, i.e., food blogs tweet about `delicious ketogenic recipe/meal preparation', lifestyle magazines tweet about `weight loss program using keto/paleo diet'.

Other users mostly retweet and share keto diet news rather than directly following the ketogenic lifestyle or promoting keto. They have following words -- \textit{rt, ketodietapp, ketogenic diet, ketone, low carb, cook, health benefit.} (Fig. \ref{fig:other_keto_wo}).

\begin{figure*}
\begin{subfigure}{.705\columnwidth}
  \centering
  \includegraphics[width=\textwidth]{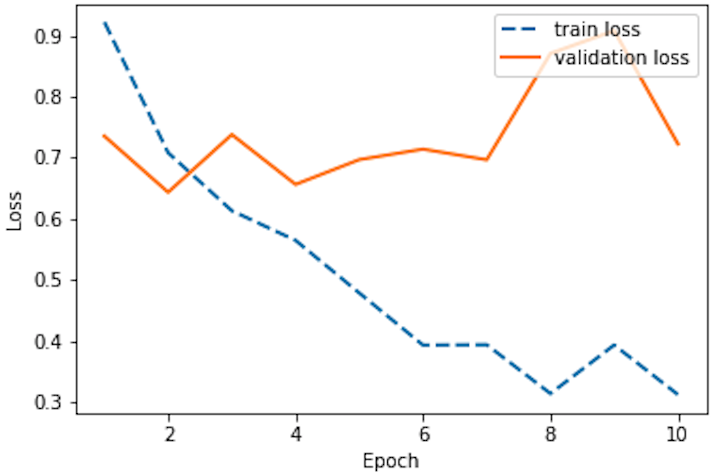}
  \caption{}
  \label{fig:yoga_loss_Bert_DLNT_@mention_utype}
\end{subfigure}%
\begin{subfigure}{.705\columnwidth}
  \centering
  \includegraphics[width=\textwidth]{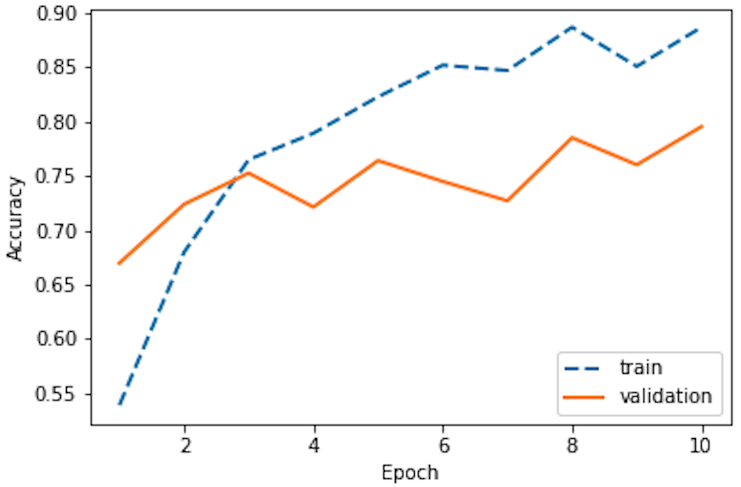}
  \caption{}
  \label{fig:yoga_acc_Bert_DLNT_@mention_utype}
\end{subfigure}
\begin{subfigure}{.705\columnwidth}
  \centering
  \includegraphics[width=\textwidth]{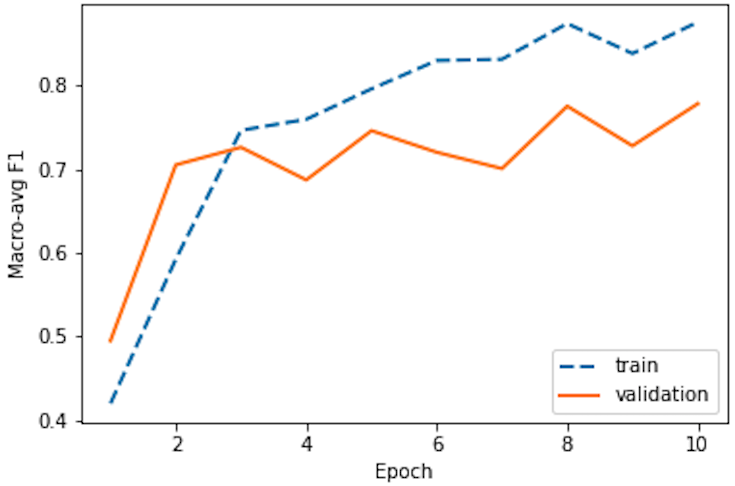}
  \caption{}
  \label{fig:yoga_f1_Bert_DLNT_@mention_utype}
\end{subfigure}
\begin{subfigure}{.705\columnwidth}
  \centering
  \includegraphics[width=\textwidth]{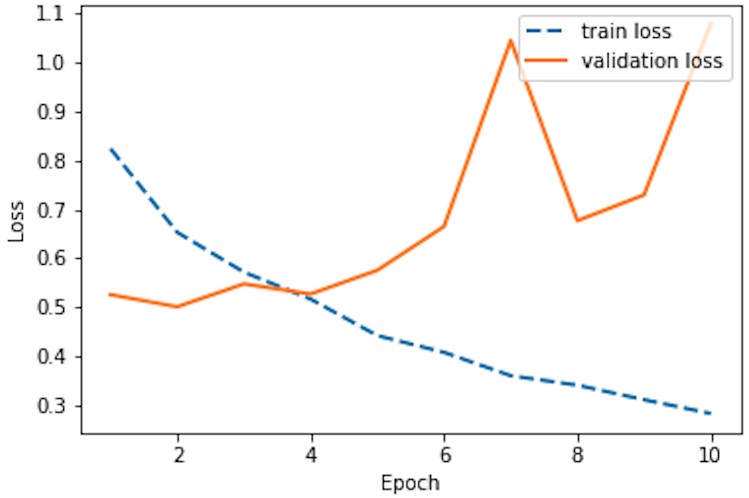}
  \caption{}
  \label{fig:yoga_loss_Bert_DLNT_@mention_umotivation}
\end{subfigure}%
\begin{subfigure}{.705\columnwidth}
  \centering
  \includegraphics[width=\textwidth]{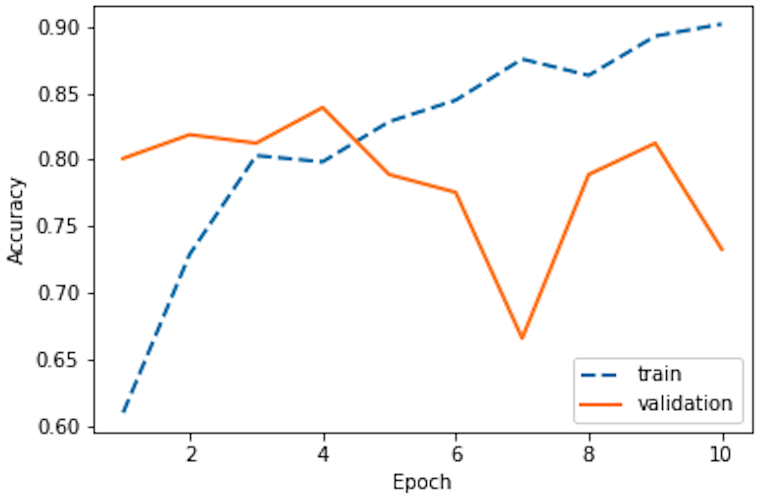}
  \caption{}
  \label{fig:yoga_acc_Bert_DLNT_@mention_umotivation}
\end{subfigure}
\begin{subfigure}{.705\columnwidth}
  \centering
  \includegraphics[width=\textwidth]{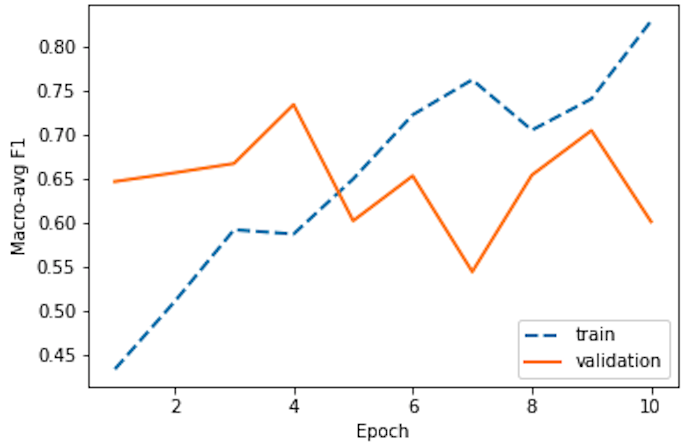}
  \caption{}
  \label{fig:yoga_f1_Bert_DLNT_@mention_umotivation}
\end{subfigure}
\begin{subfigure}{.705\columnwidth}
  \centering
  \includegraphics[width=\textwidth]{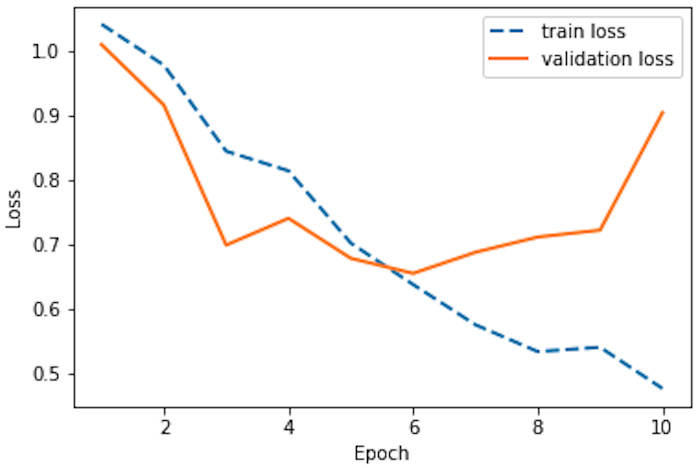}
  \caption{}
  \label{fig:keto_loss_Bert_DLNT_@mention_utype}
\end{subfigure}%
\begin{subfigure}{.705\columnwidth}
  \centering
  \includegraphics[width=\textwidth]{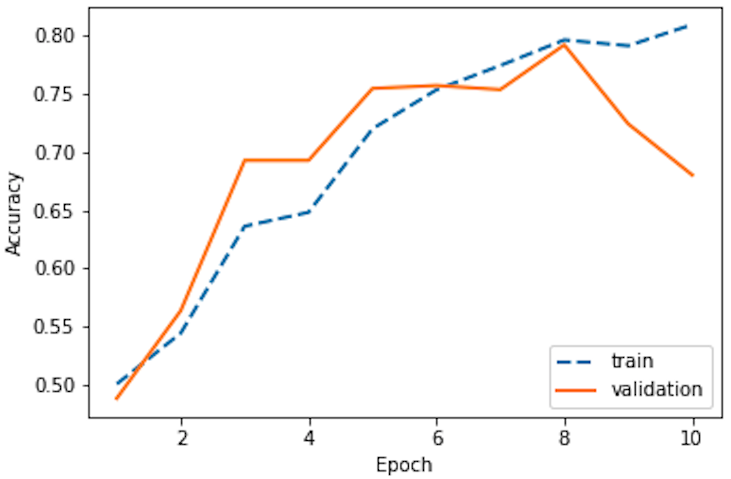}
  \caption{}
  \label{fig:keto_acc_Bert_DLNT_@mention_utype}
\end{subfigure}
\begin{subfigure}{.705\columnwidth}
  \centering
  \includegraphics[width=\textwidth]{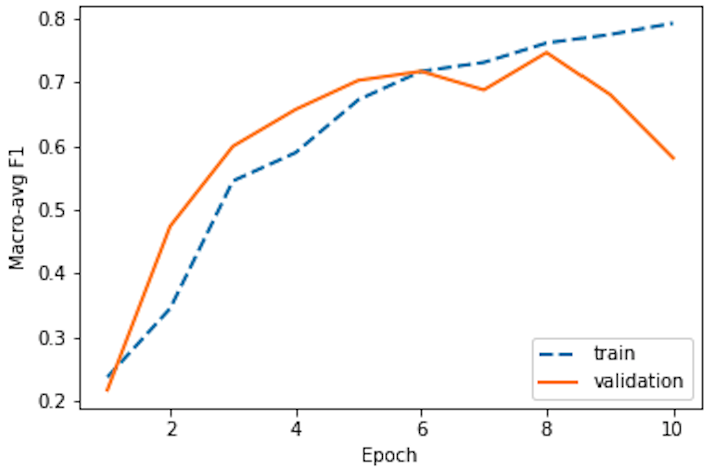}
  \caption{}
  \label{fig:keto_f1_Bert_DLNT_@mention_utype}
\end{subfigure}
\caption{Learning curves of our BERT based joint embedding model for training and validation data based on yoga and keto dataset. The blue dashed line represents train data and the orange solid line represents validation data. The $x$-axis shows number of epochs and $y$-axis corresponds to loss, accuracy, and macro-avg F1 score respectively. (a) loss vs. epochs for yoga user type, (b) accuracy vs. epochs for yoga user type, (c) macro-avg F1 score vs. epochs for yoga user type, (d) loss vs. epochs for yoga user motivation, (e) accuracy vs. epochs for yoga user motivation, (f) macro-avg F1 score vs. epochs for yoga user motivation, (g) loss vs. epochs for keto user type, (h) accuracy vs. epochs for keto user type, (i) macro-avg F1 score vs. epochs for keto user type.}
%Learning curves of our model for training and validation data based on yoga and keto data.}

\label{fig:learning_curve}
\end{figure*}

\subsection{Relationship between Descriptions and Labels}
We create wordcloud with the most frequent words (Fig. \ref{fig:yoga_keto_desc_wc}) from yoga and keto users' profile description, keeping the word `yoga' and `keto' correspondingly. We observe the words \textit{yoga, teacher, health, fitness, meditation, lover, coach, founder, author, writer, instructor, certify} in yoga practitioners' descriptions (Fig. \ref{fig:prac_yoga_desc}). Promotional users have the following words in the description \textit{yoga, fitness, wellness, community, event, offer, free, product, market, business, program, design} (Fig. \ref{fig:promo_yoga_desc}). Users who practice yoga for health benefits have similar wordcloud to yoga practitioners' descriptions (Fig. \ref{fig:health_yoga_desc}). Fig. \ref{fig:spirit_yoga_desc} shows the wordcloud of the profile description of spiritually motivated yoga user having words like \textit{yoga, spiritual, spirituality, devotee, wisdom, peace, seeker, yogi, meditator, Indian}.

Fig. \ref{fig:prac_keto_desc} and \ref{fig:promo_keto_desc} show the wordcloud of the profile description of keto practitioner having words \textit{keto, love, life, food, family} and promotional containing \textit{food, health, keto, meal, recipe, product, online, free} words respectively.

\subsection{Relationship between Location Information and Labels}
% Reviewer 1: It would be better if the authors could conduct further analysis to illustrate the relationship between location information and the label.

In this section, we illustrate the relationship between location information and user type. We use Nominatim package from GeoPy\footnote[3]{\url{https://geopy.readthedocs.io/en/stable/\#}} that given a location (either full address or city name) can identify a real-world location and provide some extra details such as latitude and longitude. To visualize the map, we use Cartopy\footnote[4]{\url{https://scitools.org.uk/cartopy/docs/v0.16/}} to plot individual locations (seen as red dots on the map), as well as blue circles whose radius varies by how many tweets come from that particular place.

In Fig. \ref{fig:all_yoga_map}, we plot yoga data distribution over the user location. 
%Fig \ref{fig:prac_yoga_map}, \ref{fig:promo_yoga_map}, and \ref{fig:other_yoga_map} show yoga practitioner, promotional, and other users data distribution over the user location. 
We observe that we have more practitioners (Fig. \ref{fig:prac_yoga_map}) and promotional users (Fig. \ref{fig:promo_yoga_map}) from the USA than the rest of the world. We find South-Asian users mostly retweet about yoga (Fig. \ref{fig:other_yoga_map}).
Fig. \ref{fig:prac_yoga_usa_map}, \ref{fig:prac_yoga_UK_map}, and \ref{fig:prac_yoga_india_map} show yoga practitioners location from USA, UK, and India. We notice more `others' users than practitioners in India (Fig. \ref{fig:other_yoga_india_map}). In Fig. \ref{fig:spirit_yoga_map}, we show yoga motivation for spirituality distribution and most of them are from India. Fig. \ref{fig:spirit_yoga_india_map} shows the yoga users from India who are motivated spiritually.

In Fig. \ref{fig:all_keto_map}, we show whole keto data distribution over the user location. 
Fig. \ref{fig:prac_keto_map}, \ref{fig:promo_keto_map}, and \ref{fig:other_keto_map} show keto practitioner, promotional, and other users data distribution over the user location. We notice that our data is skewed towards the USA. Fig. \ref{fig:prac_keto_usa_map} and \ref{fig:promo_keto_usa_map} show keto practitioners and promotional users location from the USA.
\subsection{Error Analysis}
%\subsection{Qualitative Analysis}
Overall accuracy in detecting yoga user types from our data is $85.2\%$. Our model correctly predicts $1107$ users. We have $191$ misclassifications in the yoga dataset, including $67$ misclassifications in predicting yoga practitioners, $43$ promotional users are misclassified. We notice the highest number of misclassifications ($81$) in predicting other types of users. $38$ users are misclassified as practitioners, and $43$ users are misclassified as promotional users.  

For yoga user's motivation, our model correctly predicts $1113$ users with an overall accuracy $85.6\%$. We have $185$ misclassifications, including $102$ misclassifications in predicting health-related motivation, $28$ for spiritual and $55$ other motivations are misclassified. We observe the highest number of misclassifications in predicting users' health motivation for doing yoga where $15$ and $87$ users' motivation are misclassified as spiritual and others correspondingly.

For detecting keto users, the overall accuracy is $77.6\%$. Our model correctly predicts $1009$ users. We have $291$ misclassifications in keto users, including $94$ misclassifications in predicting keto practitioners, $53$ promotional. We notice the highest number of misclassifications ($144$) in predicting other types of users. There are $96$ and $48$ users who are misclassified as practitioner and promotional, respectively.  

Our ablation study demonstrates that the profile description, tweets, and network field contribute mainly to the classification task. However, some prediction errors arise when description fields are absent or misleading. We notice that the user location has relatively low accuracy and macro-avg F1 score from our ablation study. Users sometimes do not provide location information on Twitter. Besides, as Longformer supports sequences of length up to $4096$, we might lose some information from tweets if the size of concatenated tweets $ > 4096$. Moreover, we construct $@-$mentioned network directly from retweets/mentions in tweets, which is less expensive to collect than the following network.

\section{Conclusion and Future Work}
\label{sec:6}
We propose a BERT based joint embedding model that explicitly learns contextualized user representations by leveraging users' social and textual information. We show that our model outperforms multiple baselines. Besides yoga, we demonstrate that our model can effectively predict user type on another lifestyle choice, e.g., `keto diet' 
and our approach is a general framework that can be adapted to other corpora. In the future, we aim to investigate our work to a broader impact like community detection based on different lifestyle decisions using minimal supervision.

\section*{Acknowledgments}
We are grateful to the anonymous reviewers for providing insightful reviews and Md Masudur Rahman for helping in human evaluation.

\section{Supplementary Material}
%\subsection{Additional plots}
Fig. \ref{fig:learning_curve} shows the learning curves loss (train and validation) vs. epochs, macro-avg F1 score (train and validation) vs. epochs, and accuracy (train and validation) vs. epochs for our BERT based joint embedding model on yoga and keto dataset.

\bibliography{ref}

\end{document}